\documentclass[preprint,12pt]{elsarticle}

\usepackage{amssymb}
\usepackage[T1]{fontenc}
\usepackage[table]{xcolor}
\usepackage{soul}
\usepackage{amsmath}
\usepackage{amsthm}
\usepackage{amsthm}
\newtheorem{theorem}{Theorem}   
\usepackage{multirow}
\usepackage{colortbl}
\usepackage{tabularx}
\usepackage{booktabs}

\usepackage{soul}
\usepackage{ulem}
\usepackage{subcaption}
\usepackage{booktabs}
\usepackage{graphicx,verbatim}
\usepackage{etoolbox}
\newcommand{\mSqrE}{$\square$}
\newcommand{\mSqrS}{$\square$}
\newcommand{\mSqrH}{$\square$}
\newcommand{\mTriS}{$\triangle$}
\newcommand{\mTriH}{$\triangle$}

\usepackage{graphicx}
\usepackage{amssymb}
\usepackage{soul}
\usepackage{color}
\newcommand{\lw}[1]{\color{red}#1 \color{black}}

\journal{Medical Image Analysis}

\begin{document}

\begin{frontmatter}



\title{HAPI-EP: Towards Hybrid, Adaptive, and Predictive Digital Twins of Cardiac Electrophysiology}


\author{Sumeet Vadhavkar} 
\ead{sv6234@rit.edu}
\author{Xiajun Jiang} 
\author{Yubo Ye} 
\author{Maryam Toloubidokhti} 
\author{Linwei Wang}
\affiliation{Computational Biomedicine Lab, Rochester Institute of Technology, Rochester, New York, USA}

\begin{abstract}
A
digital twin (DT) of a patient-specific heart 
offers significant potential in personalized medicine. 
However, 
its rapid and dynamic adaptation to an individual's live data 
and its predictive capability after adaptation
remains a central challenge. 
We examine this challenge from its two building blocks: 
DT formulation where mechanistic and data-driven models show competing merits and limitations, 
and DT optimization strategies that are largely driven by a reconstruction objective leading to un-identifiable models. 
We address both bottlenecks via HAPI --- an AI framework for building hybrid, adaptive, and predictive DTs with three key enablers. First, HAPI 
constructs a \textit{physics-integrated gray-box} model in which an interpretable mechanistic backbone is augmented by a neural component that models 
its \textit{residual} to
the observed data. 
Second, 
rather than attempting to pre-encode
all possible variations in a \textit{static} hybrid model, 
HAPI 
enables \textit{rapid on-the-fly 
adaptation} of the hybrid model to few-shot live data, 
achieved by feedforward meta-learners 
realizing amortized inference of both mechanistic and neural parameters of the hybrid model trained with \textit{predictive} objectives.
Finally, 
we show that this adaptivity corresponds to the construction of a conditional generative model (\textit{i.e.}, the hybrid DT) that 
endows it with theoretical identifiability and thus strong performance in predictive scenarios.  
We demonstrate the proof-of-concept of HAPI in cardiac electrophysiology using a hybrid monodomain model with mechanistic reaction kinetics and neural graph diffusion. Across synthetic and real-data studies, we show that HAPI's \textit{mechanistic-neural hybridization} and \textit{predictive adaptation} are critical for obtaining identifiable DTs with strong predictive and out-of-distribution capabilities.
\end{abstract}

\begin{graphicalabstract}
\centering
\includegraphics[width=\textwidth]{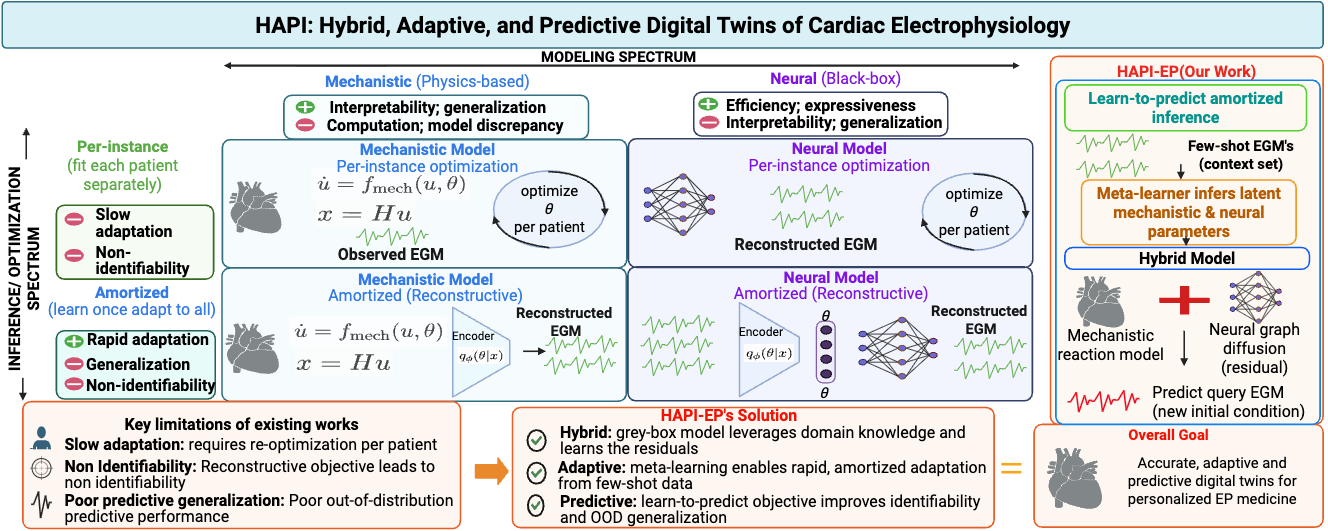}

\end{graphicalabstract}

\begin{highlights}
\item We pinpoint the challenges towards realizing the dynamic adaptation as well as predictive capability of a digital twin (DT) from its two building blocks: model formulation and optimization strategies.
\item We introduce HAPI, a novel framework for building \textit{hybrid, adaptive, and predictive} DTs, to 
enable rapid, on-the-fly adaptation of a DT through few-shot integration of live observations with 
predictiveness guaranteed by the theoretical identifiability of the adapted DT.
\item We demonstrate the proof-of-concept of HAPI in cardiac electrophysiology using a hybrid monodomain model with mechanistic reaction kinetics and neural graph diffusion, and demonstrate its strong predictive and out-of-distribution capabilities across synthetic and real-data studies.
\end{highlights}

\begin{keyword}
Digital Twin \sep Cardiac Electrophysiology \sep Hybrid and Physics-Integrated AI \sep Meta-Learning \sep Identifiability


\end{keyword}

\end{frontmatter}


\section{Introduction}
Personalized \textit{in-silico} heart models, 
such as those describing its electrophysiology (EP), have shown promises in improving personalized cardiac care such as risk stratification, diagnosis, 
and treatment planning 
\cite{Sel24,Jaffery2024,Cluitmans2024Review}. 
Towards 
their utility as a digital twin (DT) of the heart, 
there are two 
longstanding challenges. 
First, a DT needs to 
be 
dynamically and rapidly updated to the live data of an individual,
especially in 
their functional parameters. 
Second, 
once updated to the available data, 
the DT not only needs to be able to explain the observed data 
but, more crucially, to 
support decision making by predicting in scenarios beyond what are observed. 
Significant challenges exist towards these two goals, 
which we summarize below from the lens of  DT formulation 
and DT optimization strategies. 


In terms of DT formulation,
earlier works have focused on \textbf{physics-based mechanistic models} which, 
with their encoded principle of biophysics, 
are adept at extrapolation in predictive scenarios \cite{Grandits2025ECGDT,Camps2025ECGMRI_DT,Salvador2024CHD_DT,HerreroMartin2025DYNAMO,circulation2025_dt_substrate_vt,jacep2024_oversampling_dt_vt,arevalo2016arrhythmia,zahid2016feasibility}.
However, 
their computational cost raises challenges for rapid adaptation/personalization
\cite{SERMESANT2012201,wong2015velocity};  
furthermore, 
due to simplifications and imperfect knowledge, 
mechanistic models are often associated with a myriad of assumptions that cannot be resolved by optimizing model parameters alone, 
leaving significant challenges for uncertainty quantification \cite{Niederer2020VirtualCohortsUQ,Mirams2020FickleHeartUQ}.
Alternatively, machine learning (ML) and deep learning (DL) offer computationally-efficient \textbf{data-driven surrogates} to approximate the input-output relationship in mechanistic models directly from data \cite{Kashtanova2021EPNet2,morier2025learning_ce_gnn,Ziarelli2026,lydon2025physicsinformedneuraloperatorscardiac}: due to the unavailability of \textit{in-vivo} measurements of the quantities of interest, 
\textit{e.g.,} tissue properties or spatiotemporal electrical activity of the heart \cite{Cluitmans2024Review,10.1093/europace/euae295,10.1007/978-3-031-16452-1_5}, 
there has been increasing efforts in learning these surrogates from data generated from \textit{in-silico} mechanistic models. 
While these data-driven surrogates are much faster to deploy, 
they face significant challenges of generalizability and interpretability:
this includes both the simulation-to-reality gap they inherit from the \textit{in-silico} models used to train them, 
and their inherent struggle to generalize and extrapolate due to
bypassing the underlying biophysics.
\textbf{Physics-informed neural networks} (PINNs) have emerged
to bridge the gap  \cite{10.3389/fcvm.2021.768419} by 
governing the neural network with 
a mechanistic partial differential equation (PDE) representing new knowledge about the systems of interest \cite{10.3389/fcvm.2021.768419,10.1007/978-3-031-43990-2_16,10.1007/978-3-032-04927-8_33}. 
In other words, 
the mechanistic PDE \textit{informs} the PINN as a training loss which at the same time reduces data demands. 
While the resulting PINN is faster to deploy and adapt, 
it is fundamentally still a black-box that is
impacted by 
interpretability
and generalizability  
while inheriting 
imperfect knowledge in the governing PDE.

\begin{figure}[t]
    \centering
    \includegraphics[width=1\linewidth]{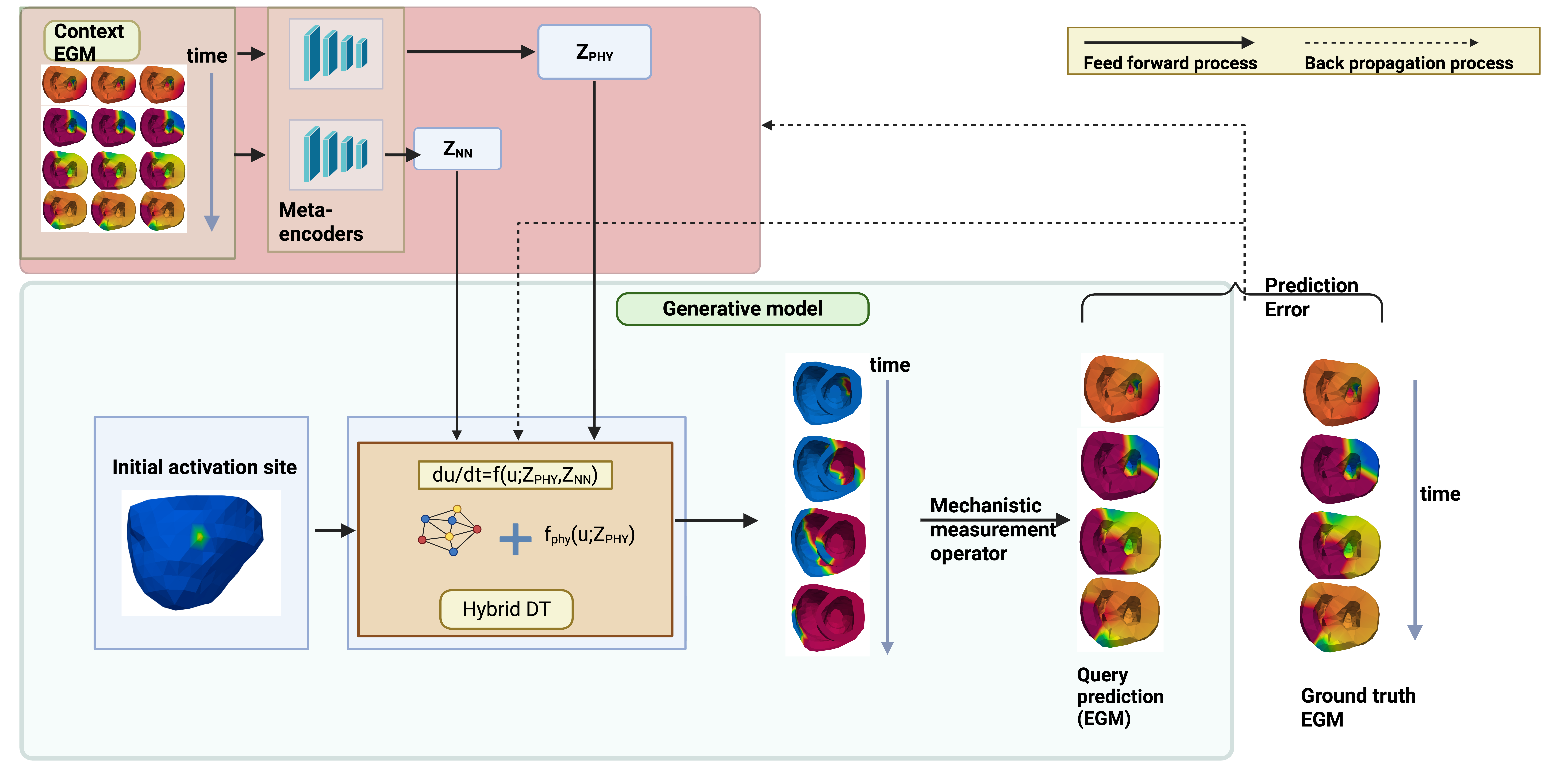}\\[2mm]

    \caption{Overview of HAPI-EP with 1) its hybrid and adaptive hybrid generative process (in light blue box), and 2) learn-to-predict meta-learning strategy for DT adaptation that ensures predictive-ness with identifiability of the DT's latent parameters (in red box).}
    \label{fig:overview}
\end{figure}

In terms of strategies to optimize the DT, 
two general approaches exist. 
Most physics- and PINN-based DTs are optimized \textit{per-instance}: 
\textit{i.e.,}
the specific parameters of the DTs are optimized to
minimize the discrepancy between the DT outputs and the specific measurements available \cite{Grandits2025ECGDT,Camps2025ECGMRI_DT,Salvador2024CHD_DT,HerreroMartin2025DYNAMO,10.3389/fcvm.2021.768419,10.1007/978-3-031-43990-2_16}. 
This often requires iterative optimization 
for adaptation 
(which we refer to as \textbf{\textit{per-instance inference}}), which
makes them 
computationally unsuitable
for clinical use. 
More recently,
amortized inference---training an \textit{inference} network to directly estimate the parameters of the DT---has emerged as a replacement 
of the traditional process of iterative optimizations, 
achieved by training across many data samples 
with which to enable
rapid adaptation at deployment time \cite{10.1007/978-3-032-04927-8_33,10.1007/978-3-031-16452-1_5}.  

However, 
current approaches 
predominantly trains the amortized inference network with a \textit{reconstruction} objective: 
\textit{i.e.}, 
the inference network is trained to infer the DT parameters from a data sample such that the adapted DT can reconstruct this input sample. 
We will show that---unless the DT is directly supervised---this reconstruction-based training objective (which we refer to as \textbf{\textit{amortized reconstructive-inference}}) can result in a non-identifiable DT with limited predictive capabilities. 
While well appreciated in mechanistic DTs \cite{Raue2009ProfileLikelihood,Wieland2021Identifiability,Chis2011StructuralIdentifiability}, 
this hurdle of identifiability and its impact on the predictiveness of DTs   
is under-recognized as neural components are increasingly appearing in DT formulations. 

In this paper, we address these fundamental roadblocks with 
HAPI, a novel framework for building \textit{hybrid, adaptive, and predictive} DTs.  
At the core of HAPI is our intention to move away from 
attempting to pre-encode all possible variations in a \textit{static} DT, 
but instead to 
enable rapid, on-the-fly adaptation of a DT through few-shot integration of live observations that directly inform and correct its residual errors, 
and to do so with predictiveness of the DT guaranteed by its identifiability. 
This is enabled by the following three novel contributions within HAPI 
outlined in Fig.~\ref{fig:overview}:
\begin{itemize}
    \item \textbf{Hybrid:} In DT formulation, we 
hybridize mechanistic and data-driven models in a  
\textit{physics-integrated} \textit{gray-box} DT, such that an interpretable and generalizable physiological model provides the central mechanistic backbone, while its residual to reality is modeled by a flexible neural component. 
This hybrid mechanistic-neural design retains interpretability and generalizability offered by our central know-hows about the mechanisms of interest, while allowing flexible data-driven correction targeting the gap in prior knowledge.  
\item \textbf{Adaptive:} In
DT optimization strategies, 
we depart from the prevalent practice 
of reconstruction-based optimization objectives which is equivalent to the optimization of a non-identifiable generative model. 
Instead,
we condition the DT on few-shot indirect observations, realized as a feedforward meta-model that is trained to 
1) use few-shot observations to identify \textit{latent} mechanistic and neural parameters of the DT, and 2) use the adapted DT to predict new data samples. 
Through this amortized \textit{learn-to-predict} meta-learner, 
we obtain a hybrid DT that can be adapted to live data via rapid feedforward computation at deployment. 

\item \textbf{Predictive:} 
We further show that this adaptivity, 
enabled by meta-learning, 
is equivalent to the construction of a conditional hybrid-DT with theoretical identifiability in its latent parameters, 
which translates to its predictiveness beyond reconstructing what is observed. 
\end{itemize}

 We demonstrated the proof-of-concept of HAPI in 
 cardiac EP, which we refer to hereafter as HAPI-EP, in 
 a hybrid monodomain model of action potential consisting of neural graph diffusion and mechanistic reaction terms. 
In synthetic and real-data experiments, 
we systematically 
evaluated the importance of 
1) the hybrid formulation 
and 2) the \textit{learn-to-predict} inference strategy 
in ensuring the 
identifiability, 
predictiveness, 
and 
out-of-distribution generalization capabilities.

\section{Background \& Related Works}
\label{sec:related}

\subsection{Background on Cardiac Electrophysiology Processes}

A common mathematical framework for modeling cardiac EP is the monodomain reaction--diffusion system, in which the transmembrane potential $u(\mathbf{r},t)$ evolves according to:
\begin{equation}
\label{eqn:mechanisticEP}
\left\{
\begin{aligned}
\chi C_m \frac{\partial u}{\partial t}
&=
\nabla \cdot \bigl( \mathbf{D} \nabla u \bigr)
-
\chi I_{\mathrm{ion}}(u,\mathbf{w})
+
I_{\mathrm{app}}, \\
\frac{d\mathbf{w}}{dt}
&=
\mathbf{F}(u,\mathbf{w}).
\end{aligned}
\right.
\end{equation}
where $\mathbf{r}$ denotes spatial locations and $t$ time. 
$C_m$ denotes membrane capacitance, $\chi$ is the membrane surface-to-volume ratio, $\mathbf{D}$ is the conductivity (or diffusion) tensor, $I_{\mathrm{ion}}$ is the total ionic current, $I_{\mathrm{app}}$ is an externally applied stimulus current, and $\mathbf{w}$ collects additional recovery, gating, and concentration variables \cite{Clayton2011,Niederer2011,Coudiere2014}. 

This formulation highlights the two core ingredients shared by a broad family of EP models. 
The diffusion term, $\nabla \cdot (\mathbf{D}\nabla u)$, governs the spatial spread of excitation through electrically coupled tissue. 
It depends on physiological properties that govern intercellular electrical coupling, including myocardial fiber organization, anisotropic tissue conductivity, and structural heterogeneity which are difficult to measure and thus associated with significant uncertainties in personalized modeling. 
The reaction terms, represented by $I_{\mathrm{ion}}(u,\mathbf{w})$ together with the state dynamics $\mathbf{F}(u,\mathbf{w})$, govern the local nonlinear membrane kinetics underlying depolarization, recovery, and refractoriness \cite{Clayton2011,Niederer2011}. 
Many models under the monodomain family differ primarily in the complexity they use to represent the reaction component: 
simplified two-variable phenomenological models, such as the Aliev--Panfilov \cite{aliev1996simple} and 
Mitchell-Schaeffer \cite{Mitchell2003} models, 
retain the essential excitable-wave and recovery behavior needed for efficient tissue-scale simulations; 
detailed biophysical ionic models, such as ten Tusscher--Panfilov \cite{tenTusscher2006} and Luo-Rudy \cite{LuoRudy1991} models, 
incorporate a much richer set of ionic currents and gating processes to more faithfully reproduce ventricular action-potential physiology.

\subsection{Existing DT Formulations and Optimization Strategies in Cardiac EP}

In this work, 
we consider DTs of cardiac EP as 
a model 
$\mathcal{M}(\mathbf{\theta})$ describing the spatiotemporal propagation of action potentials 
$\mathbf{u}_{0:T}$, 
with parameter $\theta$ that can be calibrated to patient-specific   observations $\mathbf{y}_{0:T} = g(\mathbf{u}_{0:T})$:  
once calibrated, $\mathcal{M}(\hat{\mathbf{\theta}})$ with inferred $\hat{\mathbf{\theta}}$ can be used in patient-specific simulations in predictive scenarios \cite{Grandits2025ECGDT,HerreroMartin2025DYNAMO}. 
Importantly, 
we do not consider works that are focused on inferring $\mathbf{u}_{0:T}$ or its related variables from observed $\mathbf{y}_{0:T}$ without an underlying $\mathcal{M}(\mathbf{\theta})$ that can be optimized for simulations \cite{Li2024ECGInverseSurvey,Cluitmans2017PhysiologyBasedRegularizationECGI,Wang2010TBME_TMP}.



\subsubsection{White-, Black-, and Gray-Box DT Formulations} 

\textbf{White-box mechanistic models:}
The set of mathematical models as summarized in Equation \eqref{eqn:mechanisticEP} 
represent
\textit{white-box models} 
with mathematical expressions $\mathcal{M}_{\text{PHY}}$ 
describing available biological and physiological knowledge about the EP process. 
These \textit{physics-based mechanistic DTs} represent the forefront of DT research in the domain of cardiac DT \cite{Grandits2025ECGDT,Camps2025ECGMRI_DT,HerreroMartin2025DYNAMO,circulation2025_dt_substrate_vt,jacep2024_oversampling_dt_vt,arevalo2016arrhythmia,zahid2016feasibility}. 
Their underlying physics-based principles allow these models to extrapolate in hypothetical scenarios to simulate, predict, and answer \textit{what-if} questions. 
Their white-box structures and physics-based parameters also lend interpretability to explain their simulation outputs. 
Specifications of both their structures and parameters, 
however, 
are associated with assumptions and uncertainties, 
especially for those components that are difficult to measure or 
yet resolved with scientific knowledge. 
Their computation cost also prohibits  a rapid adaptation, 
especially in the inference of their functional parameters to patient-specific values.

\textbf{Black-box neural models:}
On the other end of the spectrum, 
\textit{black-box models} describe cardiac EP process with ML or DL models $\mathcal{M}_{\phi}$ and their weight parameters $\phi$. 
These models are appreciated for their expressiveness, flexibility, and rapid computation at deployment; 
with proper design, 
these models also offer greater potential for rapid adaptation than their mechanistic counterparts. 
As any ML/DL models, 
unfortunately, 
the training of these models typically depend on a large quantity of diverse and high-quality data samples: 
while these samples are difficult to attain \textit{in-vivo}, 
many black-box models have been trained from synthetic data 
as \textit{efficient surrogates} of their mechanistic counterparts. 
These raise challenges of generalizations to distributional shifts, due to both simulation-to-reality gaps and significant physiological variabilities. 
Limited interpretability for their outputs  is another major hurdle towards their adoption as a DT for healthcare decision-making purposes.
Because of these bottlenecks, 
these \textit{blackbox neural DTs} are being actively developed but further away from translational research compared to their mechanistic counterparts, in the domain of cardiac EP \cite{10.1007/978-3-031-16452-1_5,morier2025learning_ce_gnn,Kashtanova2021EPNet2,Ziarelli2026,Naghavi2026}.

\textbf{Gray-box physics-informed models:}
To bridge the gap between the two ends of the spectrum, 
physics-informed learning has taken on a variety of forms among which the most well-known may be the 
\textit{physics-informed neural networks} (PINNs) \cite{RAISSI2019686}. 
In this type of \textit{gray-box models},
a neural-network $\mathcal{M}_\phi$ with weight parameters $\phi$, is supervised by a partial differential equation (PDE) $\mathcal{M}_{\text{PHY}}$ with known mathematical expressions and parameter $\theta$. 
When $\theta$ is known, 
the PINN can be interpreted as an alternative way of solving the PDE without traditional numerical methods. 
When $\theta$ is unknown, 
the PINN can be optimized at the same time as $\theta$, 
delivering a personalized PINN with limited data. 
By leveraging the known physics in its loss functions, 
the PINN reduces the requirement of large training data compared to its black-box counterpart. 
$\mathcal{M}_\phi$, however, 
essentially remains a black-box model that faces similar challenges of generalization and interpretability as a DT. Furthermore, it will inherit any assumptions, uncertainties, and knowledge gaps in the $\mathcal{M}_{\text{PHY}}$ used to govern its training. 
Similarly, PINN-based DTs represents an emerging line of research that has not yet reached translational studies in the domain of cardiac EP \cite{10.3389/fcvm.2021.768419,10.1007/978-3-031-43990-2_16,RuizHerrera2022FiberNet,10.1007/978-3-032-04927-8_33}.

\subsubsection{DT Optimization with Patient-Specific Measurements}

Existing approaches for optimizing the white-, black-, or gray-box DTs can be loosely categorized into two groups. 
Importantly, 
here we focus on the adaptation (\textit{i.e.}, personalization) of 
the functional parameters of 
a DT to patient-specific measurements. 
We do not include reviews of 
works on the initial construction of these DTs, 
including initial construction of a mechanistic EP model from patient-specific imaging data \cite{circulation2025_dt_substrate_vt,jacep2024_oversampling_dt_vt,arevalo2016arrhythmia,zahid2016feasibility}, or the offline training of a neural surrogate for such EP models \cite{Kashtanova2021EPNet2,morier2025learning_ce_gnn,Ziarelli2026,Naghavi2026,lydon2025physicsinformedneuraloperatorscardiac}.

\textbf{Per-instance inference:}
In most approaches, 
traditional numerical optimization strategies are employed to optimize the parameters of the DT to specific measurements from a specific instance, 
known interchangeably as \textit{personalization}, \textit{calibration}, 
or \textit{adaptation} in different works. 
These apply to the optimization of most 
mechanistic and PINN models \cite{Grandits2025ECGDT,Camps2025ECGMRI_DT,HerreroMartin2025DYNAMO,10.3389/fcvm.2021.768419,10.1007/978-3-031-43990-2_16}, 
while some recent works also considered per-instance optimization of the input mechanistic parameters $\theta$ to a  black-box neural surrogate $\mathcal{M}_{\phi}(\theta)$ to fit observed patient-specific data \cite{Naghavi2026,Salvador2024CHD_DT}. 

This \textit{per-instance inference}
can be generally expressed as: 
\begin{equation}
\label{eqn:per_instance}
    \hat{\theta} = \arg\min_{\theta} ||  g(\mathcal{M}_{\textrm{PHY}/\phi}(\mathbf{\theta})) - \mathbf{y}_{obs} ||_2^2
\end{equation}
for a mechanistic DT $\mathcal{M}_{\text{PHY}}$ 
or neural DT 
$\mathcal{M}_{\phi}$. 

In the setting of PINNs,
both $\phi$ and $\theta$ are simultaneously optimized:
\begin{equation}
    \{\hat{\phi}, \hat{\theta} \} = \arg\min_{\phi,\theta} \{ 
    ||\mathcal{M}_\phi - \mathbf{y}_{obs} ||_2^2 + 
    \lambda ||\mathcal{M}_{\text{PHY}}
    (\mathcal{M}_\phi; \theta)||_2^2 \}
\end{equation}
where the first term fits $\mathcal{M}_\phi$'s output to available data
(data-fitting loss), and 
the second term encourages $\mathcal{M}_\phi$'s output to follow the governing PDE specified by $\mathcal{M}_{\text{PHY}}$ with 
unknown physics parameter $\theta$
(PDE residual loss). 

Regardless of the underlying models, 
\textit{per-instance inference} involves iterative optimization for the purpose of DT adaptation, 
limiting their ability to rapidly adapt to limited live data. 

\textbf{Amortized reconstructive-inference:}
Alternatively, 
amortized inference represents a learning-based strategy to train a neural network (often known as the \textit{inference} or \textit{encoder} network) to infer the parameter of a DT model from available training data. 
Once trained, 
this inference network can replace iterative per-instance optimization 
with rapid feedforward computation, realizing rapid adaptation. 
In analogy to the per-instance optimization objective in Equation 
\eqref{eqn:per_instance}, 
the amortized optimization objective can be expressed as:
\begin{equation}
    \{ \hat{\omega}, \hat{\phi}
    \} = \arg\min_{\omega,\phi} ||  g(\mathcal{M}_{\textrm{PHY}/\phi}(\mathcal{E}_\omega(\mathbf{y}_{obs})) - \mathbf{y}_{obs} ||_2^2
\end{equation}
where $\mathcal{E}_\omega(\mathbf{y}_{obs})$
is the inference network parameterized by $\omega$. 
Note that the estimated parameter $\theta$ is extracted by $\mathcal{E}_\omega(\mathbf{y}_{obs})$ 
from observed $\mathbf{y}_{obs}$, 
and the adapted DT is asked to reconstruct $\mathbf{y}_{obs}$ with the estimated $\theta$. 
This is why we refer to this strategy as \textit{amortized reconstructive-inference}.
While a common concept in latent variable models such as variational auto-encoders \cite{Kingma2014VAE}, this strategy has only recently begun to appear in DT adaptation. In cardiac EP, encoder-style amortized adaptation has been explored in neural surrogate settings \cite{10.1007/978-3-031-16452-1_5}, whereas its use for mechanistic or hybrid DT adaptation remains comparatively limited \cite{10.1007/978-3-032-04927-8_33}.

\textbf{Types of measurements:} 
Another relevant point is the type of measurements used to calibrate a DT. In the current literature of cardiac EP, it is common to use indirect measures, such as noninvasive body-surface electrocardiograms (ECGs) or minimally-invasive heart-surface electrograms (EGMs), to calibrate a white-box DT \cite{Grandits2025ECGDT,Camps2025ECGMRI_DT,HerreroMartin2025DYNAMO}. 
For gray- or black-box DTs that have data-driven components, 
however, 
most existing works assume 
direct supervision on the variables (\textit{e.g.,} action potential) being modeled 
\cite{10.3389/fcvm.2021.768419,10.1007/978-3-031-43990-2_16,10.1007/978-3-032-04927-8_33,10.1007/978-3-031-16452-1_5,morier2025learning_ce_gnn,Kashtanova2021EPNet2,yin2021aphynity}: as aforementioned, this assumption is often not applicable in cardiac DTs, 
\textit{e.g.,} the spatiotemporal propagation of action potentials in the heart is only partially or indirectly observed.

\subsubsection{Summary of Related Works}

In summary, 
in the domain of cardiac EP, 
mechanistic DTs are in the most mature stage of developments towards clinical translation, 
where most are optimized per instance and thus not suitable for rapid adaptation to live data \cite{Grandits2025ECGDT,Camps2025ECGMRI_DT,Salvador2024CHD_DT,HerreroMartin2025DYNAMO}.
While neural and physics-informed alternatives are being actively pursued as efficient surrogates for mechanistic DTs, 
they are either not adaptive to patient-specific data \cite{Kashtanova2021EPNet2,morier2025learning_ce_gnn,Ziarelli2026,Naghavi2026,lydon2025physicsinformedneuraloperatorscardiac}
 or need to be optimized per instance \cite{Naghavi2026,Salvador2024CHD_DT,yin2021aphynity}. 
Although amortized inference strategies exist for such adaptations, 
they are not yet adopted for DTs of cardiac EPs and are fundamentally still based on a reconstruction objective. 
Table.~\ref{tab:dt_taxonomy} presents a structured summary of existing DT formulations and optimizations strategies, along with their representative works 
that will be considered as baselines in our experimental studies. 

\begin{table*}[t]
\centering
\caption{Taxonomy of predictive cardiac EP digital-twin (DT) formulations and adaptation strategies. Only works that infer or adapt a reusable forward model for patient-specific prediction are included.  
Works focused on initial construction of either mechanistic or neural DTs are not included. 
}
\label{tab:dt_taxonomy}
\vspace{1mm}
\footnotesize
\setlength{\tabcolsep}{4pt}
\renewcommand{\arraystretch}{1.05}
\begin{tabularx}{\linewidth}{p{2.2cm} X X}
\toprule
\textbf{Model class} & \textbf{Per-instance inference (iterative)} & \textbf{Amortized inference (learned)} \\
\midrule

\textbf{White-box} \newline (mechanistic)
&
Iterative calibration of patient-specific physical parameters in mechanistic forward models. \newline
\cite{Grandits2025ECGDT,Camps2025ECGMRI_DT,HerreroMartin2025DYNAMO,circulation2025_dt_substrate_vt,jacep2024_oversampling_dt_vt,arevalo2016arrhythmia,zahid2016feasibility}
&
Encoder-style amortized adaptation of mechanistic DT parameters; 
not yet seen in cardiac EP but exist in other physics systems \newline
\cite{ALPS}
\tabularnewline

\midrule

\textbf{Gray-box} \newline 
(PINN or hybrid)
&
Iterative personalization of network and/or physics parameters under data-fitting and PDE constraints. \newline
\cite{10.3389/fcvm.2021.768419,10.1007/978-3-031-43990-2_16}
&
Encoder-style amortized adaptation of parameters for both neural and mechanistic components; 
progress limited in cardiac EP \newline
\cite{10.1007/978-3-032-04927-8_33}
\tabularnewline

\midrule

\textbf{Black-box} \newline (neural surrogate)
&  
Iterative calibration of input mechanistic parameters  to a pre-trained neural-network surrogate \cite{Naghavi2026}\newline

&
Encoder-style amortized adaptation of parameters for neural surrogates; 
progress limited in cardiac EP \newline
\cite{10.1007/978-3-031-16452-1_5}
\tabularnewline
\bottomrule
\end{tabularx}
\end{table*}

\subsection{Existing Mechanistic-Neural Hybrid Models}

Both within and outside the domain of building DTs for cardiac EP, there has been growing interest in hybridizing physics-based models with neural augmentations \cite{NeuralSim,UDE,yin2021aphynity}. 
For instance, 
APHYNITY \cite{yin2021aphynity} augments an incomplete physics-based dynamical model with a neural residual term: 
to avoid the expressive neural component overpowers the physics-based component, 
it explicitly encourages the physics-based component to explain as much of the dynamics as possible and restricts the neural term to capture only the remaining discrepancy. 
Similarly, 
universal differential equations (UDEs) \cite{UDE} are presented as a general way to incorporate a neural network within physics-based differential equations, while 
NeuralSim \cite{NeuralSim} 
provides a broader hybrid-modeling framework in which neural modules are embedded inside differentiable simulators beyond additive formulations. All of these works, however, assume direct supervision on the state variables being modeled by the hybrid models. Furthermore, these hybrid models are commonly fit to observed trajectories via gradient-based system identification, often in a per-instance manner \cite{yin2021aphynity}.
Whether the mechanistic and neural components can be separately identified when given indirect observations, and to what extent this may affect the predictive performance of the optimized hybrid model, have not been well investigated.

Amortized inference strategies have emerged for optimizing hybrid models. 
Hybrid-VAE \cite{takeishi2021physvae} combines mechanistic dynamics with latent neural components and use an encoder to infer the latent parameters for both components from indirect observations. Robust hybrid learning \cite{Wehenkel2023RobustHybridExpertAug} improves this by 
 generating expert-augmented training examples using the approximate mechanistic component. 
These works have relaxed the assumption of direct supervision, \textit{i.e.,} latent parameters of the hybrid model are inferred from indirect observations of the variables the hybrid model describes. 
They also start to focus on the challenges in separately identifying the physics-based and neural components of the model, although from the lens of ensuring the physics-based component to not be dominated by the expressive neural component. The amortized inference fundamentally still relies on a reconstructive objective, and its impact on the identifiability of the hybrid model are not discussed in these existing works.

In a recent work, we have examined the issue of the non-identifiability of naive hybrid deep generative models under indirect observations, and established the first theoretical foundation for leveraging meta-learning as a remedy to address the issue of identifiability \cite{yubo-nps}. 
While focused on theoretical analysis and benchmarked on simple physics systems (such as pendulums), 
this works lays the theoretical basis for HAPI-EP, which we 
extend to identifiability-aware  hybrid-DTs for cardiac EP that is adaptive and predictive.



\section{Methodology}

We present HAPI-EP by its four main elements: 
the adaptive and hybrid formulation of the generative DT (Section \ref{subsec:method:generation}), 
its identifiable formulation by conditioning on context samples (Section \ref{subsec:method:conditional}), 
the corresponding \textit{learn-to-predict} inference strategy 
(Section \ref{subsec:method:inference}), 
and the conditions under which the identifiability of the hybrid DT can be established  (Section \ref{subsec:method:identifiability}).

\subsection{Hybrid and Adaptive Modeling of Cardiac EP Process}
\label{subsec:method:generation}

Let $\mathbf{u}_t\in\mathbb{R}^{N}$ denote the action potential amplitude defined over a volumetric bi-ventricular mesh represented by $N$ number of nodes at time $t$, and $\mathbf{x}_t\in\mathbb{R}^{M}$ denote the EGM amplitude measured at $M$ electrodes. 
We model $\mathbf{u}_t$ with
a hybrid dynamics process combining mechanistic reaction and neural diffusion as:
\begin{equation}
\label{eqn:hybrid}
\left\{
\begin{aligned}
\dot{\mathbf{u}}_t &= f_{\text{NN-dff}}\!\left(\mathbf{u}_t;\mathbf{z}_\text{NN}\right)
+ f_{\text{PHY-rct}}^{(u)}\!\left(\mathbf{u}_t,\mathbf{v}_t;\mathbf{z}_\text{PHY}\right)+g(\mathbf{s})
\\
\dot{\mathbf{v}}_t &= f_{\text{PHY-rct}}^{(v)}\!\left(\mathbf{u}_t,\mathbf{v}_t;\mathbf{z}_\text{PHY}\right)
\end{aligned}
\right.
\end{equation}
where 
$\mathbf{v}_t$ denotes the recovery
state at time $t$ and 
$g(\mathbf{s})$ the external stimuli at known sites $\mathbf{s}$,  \textit{e.g.}, a known pacing site. 
$f_{\textrm{PHY-rct}}$ represents a known reaction term  
and  
$f_{\textrm{NN-dff}}$ represents a diffusion term described by a light-weight 
neural network. 
This modeling choice is motivated by the fact that
subject-specific conductivity and anisotropy are difficult to measure, while explicitly resolving the diffusion term requires high-resolution numerical discretization with heavy computation
\cite{Barrios_Espinosa_2025}.
Importantly, 
neither $f_{\textrm{NN-dff}}$ nor $f_{\textrm{PHY-rct}}$ are \textit{globally} defined for all samples, 
but instead are adaptable by their respective parameters 
$
\mathbf{z}_\text{NN}$ and $
\mathbf{z}_\text{PHY}$ which we elaborate below.

\subsubsection{Mechanistic Reaction Functions}
The mechanistic terms $f_{\textrm{PHY-rct}}^{u,v}(\mathbf{u}_t,\mathbf{v}_t;\mathbf{z}_\text{PHY})$ 
describe 
nonlinear local reaction dynamics. Various models can be adopted. As a proof of concept, we consider 
the Aliev--Panfilov (AP) reaction dynamics \cite{aliev1996simple}.
Based on sensitivity analysis, we let parameter $a$ in the AP model be adaptive given its strong influence on local action potential morphology. 
The less influential parameters 
are fixed to literature values. 
In other words, 
$\mathbf{z}_\text{PHY} := a$.

\subsubsection{Neural Graph Diffusion Functions}
The neural term $f_{\text{NN-dff}}(\mathbf{u}_t;\mathbf{z}_\text{NN})$ models spatial propagation of excitation over the
bi-ventricular mesh as a learnable diffusion operator on the mesh graph.
Let the mesh be represented as a graph $\mathcal{G}=(\mathcal{V},\mathcal{E})$ with $|\mathcal{V}|=N$ nodes and edges $\mathcal{E}$ defined by $n$ nearest neighbors. 
Following \cite{grand},
we model the learned diffusion dynamics as
\begin{equation}
f_{\text{NN-dff}}(\mathbf{u}_t;\mathbf{z}_\text{NN})
=
\alpha \, \bigl(\mathbf{A}(\mathbf{u}_t)-\mathbf{I}\bigr)\,\mathbf{u}_t
\;\triangleq\;
\alpha\,\bar{\mathbf{A}}(\mathbf{u}_t)\,\mathbf{u}_t,
\label{eq:grand_u_ode}
\end{equation}
where $\mathbf{A}(\mathbf{u}_t)$ is an adjacency-structured attention operator with support on
$\mathcal{E}$ (\textit{i.e.}, nonzero only on $n$ nearest neighbors). 
We adopt \textit{scaled dot-product attention} as:
\[
A(u_{t,i},u_{t,j})=\mathrm{softmax}_{j\in\mathcal{N}(i)}
\Bigl(\tfrac{(\mathbf{W}_Q\,u_{t,i})^\top (\mathbf{W}_K\,u_{t,j})}{\sqrt{d}}\Bigr),
\qquad (i,j)\in\mathcal{E},
\]
where 
$d$ is the dimensionality of the query/key embeddings.
$\mathbf{W}_Q$ and $\mathbf{W}_K$ are expected to learn patient-specific anisotropic conduction while $\alpha$ modulates 
conduction velocity. 
We thus define $\mathbf{z}_\text{NN} := \{ \alpha, \mathbf{W}_Q, \mathbf{W}_K \}$ to be adaptive.

\subsubsection{Mechanistic Emission/Measurement Processes}

We describe the measurement process for $\mathbf{u}_t$ with a mechanistic 
operator $\mathbf{H}$, which is  obtained by solving the Poisson equation under the
quasi-static approximation between 
the cardiac source due to $\mathbf{u}_t$ and its extracellular potential measurements $\mathbf{x}_t$ on an surface enclosing the ventricles 
\cite{Wang2010TBME_TMP} as 

\begin{equation}
\label{eqn:emission}
    \mathbf{x}_t = \mathbf{H}\,\mathbf{u}_t + \epsilon 
\end{equation}
where $\epsilon$ is assumed to be Gaussian.  In this paper, we consider $\mathbf{x}_t$'s as electrogram (EGM) measurements on the bi-ventricular surface (synthetic data experiments) or epicardial surface only (real data experiments).

\subsubsection{Non-Identifiability of Unconditional Generative Models}
Given a known stimulation protocol $\mathbf{s}$, the hybrid EP model in
\eqref{eqn:hybrid} induces a distribution over the observed EGM sequence
$\mathbf{x}_{0:T}=\{\mathbf{x}(t)\}_{t=0}^{T}$ through latent EP dynamics.
The marginal likelihood of $\mathbf{x}_{0:T}$ is obtained by integrating out these latent variables as: 
\begin{eqnarray}
 \label{eq:hadit_likelihood_marginal}
p(\mathbf{x}_{0:T}\mid \mathbf{s})
& =& 
\int \!\! \int
p(\mathbf{x}_{0:T}\mid \mathbf{s},\mathbf{z}_\text{PHY},\mathbf{z}_\text{NN})\,p(\mathbf{z}_\text{PHY})\,p(\mathbf{z}_\text{NN}) \, d\mathbf{z}_\text{PHY}\, d\mathbf{z}_\text{NN}
\end{eqnarray}
where 
\begin{eqnarray}
\small{p(\mathbf{x}_{0:T}\mid \mathbf{s},\mathbf{z}_\text{PHY},\mathbf{z}_\text{NN})  = 
\prod_{t=0}^{T}
p\!\left(\mathbf{x}_t\mid \mathbf{u}_t\right)|_{\mathbf{u}_t  =  \mathbf{u}_{t-1}+\int_{s=t-1}^t f_\text{hyd}(\mathbf{u}_s, \mathbf{v}_s; \mathbf{z}_{\text{PHY}},\mathbf{z}_{\text{NN}} )ds, \mathbf{u}_0 = g(\mathbf{s})}}
 \end{eqnarray}
where $f_\text{hyd}(\mathbf{u}_s, \mathbf{v}_s; \mathbf{z}_{\text{PHY}},\mathbf{z}_{\text{NN}} )$ describes the hybrid model in Equation \eqref{eqn:hybrid} and 
$p(\mathbf{x}_{t}\mid \mathbf{z}_t)$ is defined by the emission Equation \eqref{eqn:emission}.

Existing works in DT optimization are mostly based on the maximization of this likelihood, 
with learnable parameters 
(\textit{e.g.}, $\mathbf{z}_\text{PHY}$ and $\mathbf{z}_\text{NN}$)
inferred from $\mathbf{x}_{0:T}$. 
This corresponds to the inference of a generative model whose latent variables 
($\mathbf{z}_\text{PHY}$ and $\mathbf{z}_\text{NN}$) have an unconditional prior.
As established by the theory of nonlinear independent component analysis (ICA) \cite{khemakhem2020ivae}, 
the neural latent variable $\mathbf{z}_\text{NN}$  of such an unconditional generative model 
is not identifiable \cite{yubo-nps}: 
while we refer the audience to 
\cite{khemakhem2020ivae,yubo-nps} for theoretical details, 
intuitively it means that an arbitrary nonlinear transformation can change the value of $\mathbf{z}_\text{NN}$ without changing its distribution, 
and this transformation can be absorbed by the neural function without affecting the decoding to $\mathbf{x}_{0:T}$. 
Furthermore, 
this lack of identifiability in $\mathbf{z}_\text{NN}$
may then compromise the recovery of $\mathbf{z}_\text{PHY}$, 
\textit{i.e.}, multiple values
of $\mathbf{z}_\text{PHY}$ and $\mathbf{z}_\text{NN}$ 
may produce observationally
equivalent $p(\mathbf{x}_{0:T}\mid \mathbf{s})$ \cite{yubo-nps}: 
while these models may perform similarly in reconstructing $\mathbf{x}_{0:T}$, 
their ability to predict beyond $\mathbf{x}_{0:T}$ may be impacted .

\subsection{Identifiable Formulation via Context-Conditional Priors}
\label{subsec:method:conditional}

To construct an identifiable hybrid DT, we draw inspiration from \cite{yubo-nps} and condition the generation of $\mathbf{x}_{0:T}$ on \textit{auxiliary information} defined as 
$k$-shot examples $\mathcal{D}^c$ that are distinct from $\mathbf{x}_{0:T}$ but share its generative parameters $\mathbf{z}_\text{PHY}$ and $\mathbf{z}_\text{NN}$.
More specifically, 
this is achieved by replacing the unconditional prior of $\mathbf{z}_\text{PHY}$ and $\mathbf{z}_\text{NN}$ with their conditional counterparts $p(\mathbf{z}_\text{PHY} \mid \mathcal{D}^c)$ and $
p(\mathbf{z}_\text{NN} \mid \mathcal{D}^c)$.

\subsubsection{Context-Conditioned  Priors for the Latent Hybrid Parameters} 
With $\mathbf{z}_\text{PHY} := a$, 
we parameterize the conditional prior of the mechanistic reaction parameter as:
\begin{equation}
\mathbf{h}^{(a)}_{i} \sim \mathcal{N} (E_{\phi_1}(\mathbf{x}^{c}_i), \sigma)
\qquad
\hat{a} = \text{Agg}\big( \{ g_{\phi_2}\!\big(\mathbf{h}^{(a)}_i\big) \}_{\mathbf{x}_i^c \in \mathcal{D}^c}\big)
\label{eq:cond_prior_a_dirac}
\end{equation}
where $E_{\phi}$ and $g_\phi$  are neural networks parameterized by $\phi_1$ and $\phi_2$, respectively. 
They are applied to each context sample $\mathbf{x}_i^c \in \mathcal{D}^c$ to extract a scalar embedding, and 
$\mathrm{Agg}(\cdot)$ denotes mean pooling across the context set.
In our implementation,
$E^{(a)}_{\phi_1}(\cdot)$ is a spatio-temporal ResNet-GCN with temporal 1D convolutions followed by a GCNConv over mesh edges and global pooling. $g_{\phi_2}$ is a lightweight MLP head. 
In other words, 
$\hat{a}$ is an embedding shared by all samples in the context set, 
which remove sample-specific information (\textit{e.g.}, those related to initial conditions) 
and extract shared parameters from available context samples. 
In this proof of concept study, 
we consider $\sigma$ to be a pre-defined constant.

Similarly, with $\mathbf{z}_\text{NN} := \{ \alpha, \mathbf{W}_Q, \mathbf{W}_K \}$, 
we define the conditional prior for the neural diffusion parameters as:
\begin{eqnarray}
\nonumber
\{ \mathbf{r}^{(Q)}_i,\ \mathbf{r}^{(K)}_i,\ \hat{\alpha}_i \}
&= &
\mathcal{N}(
E_{\psi_1}(\mathbf{x}^c_i), \sigma),
\\
\nonumber
\{\bar{\mathbf{r}}^{(Q)},\bar{\mathbf{r}}^{(K)},  
\hat{\alpha} \}
&=&\mathrm{Agg}\big(\{\mathbf{r}^{(Q)}_i, \mathbf{r}^{(K)}_i, \hat{\alpha}_i\}_{\mathbf{x}_i^c \in \mathcal{D}^c}\big)
\\
\mathbf{W}_Q  =  \mathrm{HNN}_{\psi_2}\!\left(\bar{\mathbf{r}}^{(Q)}\right),
&\quad&
\mathbf{W}_K = \mathrm{HNN}_{\psi_2}\!\left(\bar{\mathbf{r}}^{(K)}\right),
\label{eq:hnn_wq_wk}
\end{eqnarray}
where $E_{\psi_1}(\cdot)$ uses the same ResNet--GCN backbone as $E_{\phi_1}$, followed by three lightweight MLP heads. $\mathrm{HNN}_{\psi_2}$ is a shared hyper-network parameterized by $\psi_2$. 
In other words, 
$\bar{\mathbf{r}}^{(Q)},\bar{\mathbf{r}}^{(K)}, \ 
\hat{\alpha}$ extract embeddings shared by all context samples, 
where $\hat{\alpha}$ modulates the neural diffusion strength in Equation \eqref{eq:grand_u_ode}, 
and $\{\bar{\mathbf{r}}^{(Q)}, \bar{\mathbf{r}}^{(K)}\}$ adapt the projection matrices in the attention-based diffusion operator in Equation \eqref{eq:grand_u_ode}.

\subsubsection{Conditional Generative Model} 

The context-conditioned likelihood of $\mathbf{x}_{0:T}$ is now:
\begin{equation}
\small{
p(\mathbf{x}_{0:T} \mid \mathbf{s}, \mathcal{D}^c)
=
\int\!\!\int
p(\mathbf{x}_{0:T}\mid \mathbf{s},\mathbf{z}_\text{PHY},\mathbf{z}_\text{NN})\,
p_\phi(\mathbf{z}_\text{PHY} \mid \mathcal{D}^c)\,p_\psi(\mathbf{z}_\text{NN}\mid \mathcal{D}^c)
\,d\mathbf{z}_\text{PHY}\,d\mathbf{z}_\text{NN}}
\label{eq:hadit_conditional_likelihood}
\end{equation}
where 
$\phi = \{\phi_1, \phi_2\}$ as defined in Equation \eqref{eq:cond_prior_a_dirac} and 
$\psi = \{\psi_1, \psi_2\}$ as defined in Equation \eqref{eq:hnn_wq_wk}. 
The generation of  $\mathbf{x}_{0:T}$ is now conditioned on 
context samples $\mathcal{D}^c$ that are distinct from $\mathbf{x}_{0:T}$ but share its latent parameters. 
$\mathcal{D}^c$
adapts the distribution of 
$\mathbf{z}_\text{PHY}$
and $\mathbf{z}_\text{NN}$ as defined by 
Equations \eqref{eq:cond_prior_a_dirac}-\eqref{eq:hnn_wq_wk}, 
via which they adapt 
the hybrid DT to be specifically predictive for $\mathbf{x}_{0:T}$: we refer to $\mathbf{x}_{0:T}$ as \textit{query} samples hereafter and denote it as $\mathbf{x}^q$ for clarity.

\noindent

\subsection{Amortized Variational Inference as Meta-Learning}
\label{subsec:method:inference}

To maximize the conditional likelihood as defined in  
Equation \eqref{eq:hadit_conditional_likelihood},  we introduce the variational approximation of 
the posterior densities 
of 
$\mathbf{z}_\text{PHY}$ and $\mathbf{z}_\text{NN}$ as
$q_\phi(\mathbf{z}_\text{PHY} \mid \mathcal{D}^c \cup \mathbf{x}^q)$
and 
$q_\psi(\mathbf{z}_\text{NN} \mid \mathcal{D}^c \cup \mathbf{x}^q)$, 
sharing the same networks defining their priors as in Equations \eqref{eq:cond_prior_a_dirac}-\eqref{eq:hnn_wq_wk}.
This gives rise to the following evidence lower bound ($\mathcal{L}_{\mathrm{ELBO}}$) of 
$\log p(\mathbf{x}^q \mid \mathbf{s}^q, \mathcal{D}^c)$ as: 
\begin{align}
\log p(\mathbf{x}^q \mid \mathbf{s}^q, \mathcal{D}^c)\ \ge\ 
\mathcal{L}_{\mathrm{ELBO}}
&= \mathbb{E}_{q_{\phi}(\mathbf{z}_{\text{PHY}})
q_\psi(\mathbf{z}_{\text{NN}})}
\Big[\log p(\mathbf{x}^q \mid \mathbf{s}^q, \mathbf{z}_\text{PHY},\mathbf{z}_\text{NN})\Big]
\nonumber \\\nonumber
&\quad
-\mathrm{KL}\!\left(q_{\phi}(\mathbf{z}_\text{PHY}\mid \mathcal{D}^c \cup \mathbf{x}^q )\ \|\ p_{\phi}(\mathbf{z}_\text{PHY}\mid \mathcal{D}^c)\right)\\
&-\mathrm{KL}\!\left(q_{\psi}(\mathbf{z}_\text{NN}\mid \mathcal{D}^c \cup \mathbf{x}^q)\ \|\ p_{\psi}(\mathbf{z}_\text{NN}\mid \mathcal{D}^c)\right).
\label{eq:hadit_elbo_prediction}
\end{align}
where $q_{\phi}(\mathbf{z}_{\text{PHY}})$ and 
$q_\psi(\mathbf{z}_{\text{NN}})$ 
are short for 
$q_\phi(\mathbf{z}_\text{PHY} \mid \mathcal{D}^c \cup \mathbf{x}^q)$
and 
$q_\psi(\mathbf{z}_\text{NN} \mid \mathcal{D}^c \cup \mathbf{x}^q)$, respectively. 
Note that the first likelihood term now defines a
\textit{predictive} objective 
where the hybrid DT is asked to use the latent parameters $\mathbf{z}_\text{PHY}$ and $\mathbf{z}_\text{NN}$,  
inferred from context samples $\mathcal{D}^c$, 
to predict for query examples $\mathbf{x}^q$ that are 
different from $\mathcal{D}^c$.
This is fundamentally different from reconstruction-based objectives.

Interestingly, optimizing the ELBO in~\eqref{eq:hadit_elbo_prediction} corresponds to a feedforward meta-learning process
performed over \textit{tasks} defined by 
individual patients or patient-subgroups, depending on the problem of interest. 
Given data from multiple such tasks,
the conditioning on $\mathcal{D}^c$ realized via Equations \eqref{eq:cond_prior_a_dirac}-\eqref{eq:hnn_wq_wk} 
work as meta-models that provide inner-loop task adaptation, 
and 
the expected query log-likelihood defines the 
outer-loop meta-objective for optimizing  the meta-models.
Notably, 
unlike gradient-based meta-learning (\textit{e.g.}, model-agnostic meta-learning (MAML) \cite{maml}) where the inner-loop adaptation performs explicit gradient steps,
HAPI-EP performs amortized adaptation in the form of feedforward passes through meta-models 
$E_{\phi_1},g_{\phi_2}, E_{\psi_1}$ and $\mathrm{HNN}_{\psi_2}$.

\subsection{Identifiability Theory}
\label{subsec:method:identifiability}

Built on the theory established in \cite{yubo-nps}, 
we can establish the identifiability of the latent neural parameter $\mathbf{z}_\text{NN}$ of the conditional hybrid DT.
\vspace{-.1cm}
    \begin{theorem}
    \label{theorem:mcc}
    Assume that we observed data 
    $p(\mathbf{x}_{0:T})$ generated 
    following the process defined by Equations \eqref{eq:cond_prior_a_dirac}-\eqref{eq:hadit_conditional_likelihood} 
    with parameters 
        $(\phi, \psi )$ and 
        latent variable $p(\mathbf{z}_\text{NN}|\mathcal{D}^c)$ 
        defined as in Equation \eqref{eq:hnn_wq_wk}. 
        Denote this generation process as 
        $\mathbf{x}_{0:T} = \mathcal{G}(\mathbf{z}_\text{NN})$.
         Assume a model with parameter $(\hat{\phi}, \hat{\psi})$ and $p(\hat{\mathbf{z}}_\text{NN}|\mathcal{D}^c)$ 
         generates 
         $p(\hat{\mathbf{x}}_{0:T})$ 
         that is observationally equivalent to   $p(\mathbf{x}_{0:T})$. 
        Assume the following hold:
        \vspace{-.2cm}
        \begin{enumerate}
            \itemsep 0em
            \item The set $\left \{ \mathbf{x}_{0:T} \in \mathcal{X} | \varphi_\varepsilon(\mathbf{x}_{0:T})=0 \right \}$ has measure zero, where $\varphi_\varepsilon$ is the characteristic function of the density $p_\varepsilon(\mathbf{x}_t - \mathbf{Hu}_t).$
            \item The function $\mathcal{G}$ is injective. 
            \item There exist $de+1$ distinct context sets $\mathcal{D}^{s,0}, \mathcal{D}^{s,1},\dots, \mathcal{D}^{s,de}$ such that the $de \times de$ matrix $\mathbf{L}$ defined as follows is invertible
            \begin{equation}
            \mathbf{L} = (\pmb{\lambda}(\mathcal{D}^{c,1})-\pmb{\lambda}(\mathcal{D}^{c,0}), \dots, \pmb{\lambda}(\mathcal{D}^{c,de})-\pmb{\lambda}(\mathcal{D}^{c,0}))
        \end{equation}
        where 
        $\pmb{\lambda}$ denotes the parameters of 
        $p(\mathbf{z}_\text{NN}|\mathcal{D}^c)$, $d$ the dimension of $\mathbf{z}_\text{NN}$, and $e$ the dimension of its  sufficient statistics $\mathbf{T}$. \vspace{-.2cm}
        \end{enumerate}
       then there exists an  invertible $de\times de$ matrix $\mathbf{A}$ and vector $\mathbf{b}$, such that $\mathbf{T}(\mathbf{z}_\text{NN}) = \mathbf{A}\mathbf{T}(\hat{\mathbf{z}}_\text{NN}) + \mathbf{b}$, 
    \textit{i.e.}, $\mathbf{T}(\mathbf{z}_\text{NN})$ are identifiable up to affine transformations. 
    \end{theorem}

The proof of Theorem 1 directly builds on that 
in \cite{yubo-nps}. Interestingly, condition 3 in Theorem 1 specifies that the recovery of the latent variable $\mathbf{z}_\text{NN}$ requires sufficient variability of its values, as conditioned on $\mathcal{D}^c$, to be observed. 
In other words, for a Gaussian distribution used in this paper, a minimum of $2d+1$ unique values of $\mathbf{z}_\text{NN}$ need to be present in the observed data in order for it to be identified.

\section{Synthetic Data Experiments}

\label{subsec:exp:synth_data}

We first evaluated HAPI-EP trained and tested on synthetic data generated on different bi-ventricular models.

\subsection{Experimental Settings}

\textbf{Experimental data:} 
We considered three human bi-ventricular models derived from computed tomographic scans. 
On each mesh, 
we generated synthetic ventricular action potential sequences using the Aliev-Panfilov model \cite{aliev1996simple} with mechanistic diffusion \cite{Wang2010TBME_TMP}. 
Synthetic observations of EGMs 
were simulated on the ventricular surface using the physics-based forward operator 
numerically solved as described in 
\cite{Wang2010TBME_TMP}. 
We varied reaction parameters
$a\in\{0.08,0.10,0.12,0.14\}$ to 
control local action potential dynamics, along with diffusion parameter
$D\in\{1.5,2.0,2.5\}$ to control conduction velocity across the mesh. 
On each mesh under various combinations of $\{a, D\}$ values, 
we generated action potential simulations 
with $\sim 800$ different initial conditions uniformly distributed across the ventricular volume.
This resulted in $\sim 2,376$ samples which we split into train/val/test sets with a 60:20:20 ratio.

\textbf{Baselines:} 
We compared HAPI-EP to the spectrum of baselines summarized in Table \ref{tab:dt_taxonomy}. 
This includes
three families of model formulations:
(i) fully neural, 
(ii) hybrid, and
(iii) full physics, 
with each adopting one of the two inference strategies: 
per-instance inference, or amortized
reconstructive-inference. 
For physics-based baselines, 
we considered the full AP model 
with its parameters 
$(a,D)$ optimized 
per instance via the derivative-free BOBYQA method \cite{powell2009bobyqa}, 
or with an encoder that resembles the approach presented in ALPS \cite{ALPS}. 
For the hybrid baselines, 
its per-instance optimization is represented by APHYNITY \cite{yin2021aphynity}, 
and its amortized inference resembles 
the hybrid-VAE (H-VAE) \cite{takeishi2021physvae}.
For the 
neural baselines, we considered DeepCardioSim (specifically the GNN option presented)  \cite{Naghavi2026} for per instance optimization. We first pre-trained DeepCardioSim as a neural surrogate on synthetic EP data: following the published model architecture, it 
took as input the heart geometry, fiber directions, pacing-location encoding, and the EP parameters $(a,\alpha)$, 
and it outputs activation time (AT) on the ventricular mesh. 
The trained DT was then adapted/personalized per instance to available AT measurements, 
by freezing the network weights and optimizing only $(a,\alpha)$ using derivative-free BOBYQA method. 
Amortized inference for neural EP models did not seem to exist in published works
(although a representative version of it will be included in our ablation studies in Section \ref{subsec:ablations}).

\textbf{Evaluation scenarios \& metrics:} 
For all models, 
we evaluated their test-time performance in two settings. 
In \textit{reconstruction} tests, 
the model was asked to infer 
latent parameters from a sample and 
use them to reconstruct the same sample.
In \textit{prediction} tests, 
the model was asked to infer 
latent parameters from a sample, 
but use them to predict another sample with the same parameters but different initial conditions. 
We considered the above evaluations considering latent parameter values both within the train distribution 
and out-of-distribution (OOD).

For reconstructed/predicted action potential, 
we measured its mean squared error (MSE) with ground-truth signals over time and space, 
as well as the mean absolute error (MAE) for the activation time (AT), repolarization (RT), 
and action potential duration (APD) extracted from it. 
For physics-based parameters $a$, 
we reported the MSE between estimated and true parameter values. 
For latent neural diffusion parameters, 
we measured its mean correlation coefficients (MCC), a common identifiability metric, with true parameters \cite{khemakhem2020ivae}: the MCC value is between 0 and 1, where a higher MCC value indicates better identifiability results. 


\subsection{Results}

\subsubsection{Reconstruction \textit{versus} Prediction}

\begin{table}[t]
\centering
\caption{Summary of quantitative results on reconstruction and prediction performances of HAPI-EP and all baselines. DeepCardioSim, BoBYQA, and APHYNITY represent neural, mechanistic, and hybrid DTs with per-instance optimization (\mSqrH). ALPS and H-VAE represent mechanistic and hybrid DTs with amortized reconstructive-inference (\mTriH). HAPI-EP is hybrid with amortized predictive inference. 
Lower is better for MSE/MAE, and higher is better for MCC.} 
\label{tab:main_results_ep_combined}
\resizebox{\linewidth}{!}{
\begin{tabular}{lccc|ccc}
\toprule
\multicolumn{7}{c}{\textbf{Reconstruction}} \\
\midrule
\textbf{Method}
& \textbf{MSE($z_{\mathrm{phy}}$)}
& \textbf{MCC($z_{\mathrm{NN}}$)}
& \textbf{MSE(recon)}
& \textbf{MAE(APD)}
& \textbf{MAE(AT)}
& \textbf{MAE(RT)} \\
\midrule
DeepCardioSim \mSqrE \cite{Naghavi2026}
& -- & -- & -- & -- & 3.01 $\pm$ 1.37 & -- \\
BoBYQA \mSqrS \cite{powell2009bobyqa}
& 5.00e-4 $\pm$ 6.00e-4 & 0.250 & 0.1000 $\pm$ 0.0890 & 14.72 $\pm$ 12.09 & 6.58 $\pm$ 3.11 & 13.60 $\pm$ 11.09 \\

APHYNITY \mSqrH \cite{yin2021aphynity}
& 2.00e-4 $\pm$ 2.00e-4 & 0.310 & 0.0680 $\pm$ 0.0300 & 13.21 $\pm$ 10.48 & 5.40 $\pm$ 1.74 & 13.20 $\pm$ 11.13 \\
ALPS \mTriS \cite{ALPS}
& 7.59e-5 $\pm$ 9.50e-5 & 0.371 & \textbf{0.0124 $\pm$ 0.0146} & 2.69 $\pm$ 2.60 & \textbf{0.73 $\pm$ 0.86} & 2.68 $\pm$ 2.60 \\
H-VAE \mTriH \cite{takeishi2021physvae}
& 8.50e-5 $\pm$ 1.00e-4 & 0.340 & 0.0250 $\pm$ 0.0152 & 3.70 $\pm$ 3.12 & 1.90 $\pm$ 1.13 & 3.64 $\pm$ 2.41 \\
HAPI-EP
& \textbf{1.73e-5 $\pm$ 2.77e-5} & \textbf{0.714} & 0.0188 $\pm$ 0.0123 & \textbf{1.76 $\pm$ 1.31} & 1.88 $\pm$ 1.31 & \textbf{2.09 $\pm$ 0.95} \\
\midrule
\multicolumn{7}{c}{\textbf{Prediction}} \\
\midrule
\textbf{Method}
& \textbf{MSE($z_{\mathrm{phy}}$)}
& \textbf{MCC($z_{\mathrm{NN}}$)}
& \textbf{MSE(pred)}
& \textbf{MAE(APD)}
& \textbf{MAE(AT)}
& \textbf{MAE(RT)} \\
\midrule
DeepCardioSim \mSqrE \cite{Naghavi2026}
& -- & -- & -- & -- & 3.42 $\pm$ 1.71 & -- \\
BoBYQA \mSqrS \cite{powell2009bobyqa}
& 5.00e-4 $\pm$ 6.00e-4 & 0.300 & 0.1000 $\pm$ 0.0900 & 15.52 $\pm$ 12.38 & 5.84 $\pm$ 2.82 & 15.00 $\pm$ 11.77 \\

APHYNITY \mSqrH \cite{yin2021aphynity}
& 4.10e-4 $\pm$ 2.51e-4 & 0.320 & 0.0720 $\pm$ 0.0500 & 13.22 $\pm$ 10.48 & 5.60 $\pm$ 2.54 & 13.39 $\pm$ 10.26 \\
ALPS \mTriS \cite{ALPS}
& 9.17e-5 $\pm$ 2.20e-4 & 0.300 & \textbf{0.0168 $\pm$ 0.0220} & 3.60 $\pm$ 4.69 & \textbf{0.80 $\pm$ 0.89} & 3.60 $\pm$ 4.49 \\
H-VAE \mTriH \cite{takeishi2021physvae}
& 2.00e-4 $\pm$ 4.0e-4 & 0.259 & 0.0347 $\pm$ 0.0340 & 5.30 $\pm$ 7.33 & 2.20 $\pm$ 1.33 & 5.27 $\pm$ 6.45 \\
HAPI-EP
& \textbf{1.74e-5 $\pm$ 2.57e-5} & \textbf{0.702} & 0.0196 $\pm$ 0.0134 & \textbf{1.78 $\pm$ 1.31} & 1.94 $\pm$ 1.33 & \textbf{2.19 $\pm$ 1.12} \\
\bottomrule
\end{tabular}
}
\end{table}

The top half of 
Table~\ref{tab:main_results_ep_combined} summarizes the performance of HAPI-EP and all considered baselines 
in reconstructing the sample used to infer the latent parameters (reconstruction), 
\textit{versus} using the inferred parameters to predict for a different sample that shares the same parameters but has a different initial condition 
(prediction).
Comparing DT performance in reconstruction \textit{vs}.\ prediction settings allow us to assess not only how well a DT adapts to available measurements and to what extent the latent DT parameters are recovered during the adaptation, but also how well the adapted DT is able to predict beyond what is observed.

In reconstruction performance, 
the three per-instance models delivered weaker performances compared to the rest of the amortized models. 
Note that although
DeepCardioSim delivered a lower MAE for the estimated AT than its per-instance mechanistic and hybrid counterparts, 
it was directly optimized from measured AT and had a narrower task of producing only AT
(due to its model formulation). In comparison, 
the other models had a more challenging task of reconstructing more complete physiological variables from indirect EGM observations. 
These results revealed the potential limitations with per-instance DT optimization regardless of their underlying formulations. 

Interestingly, 
the use of amortized inference substantially improved the performance of the DT --- both mechanistic (ALPS) and hybrid (H-VAE) --- compared to their per-instance optimized counterparts, especially in the accuracy of the reconstructed action potential (MSE) and its derived physiological variables (MAE of AT, RT, and APD). 
Further note that H-VAE had similar accuracy in the estimation of reaction parameter $a$ with BoBYQA and APHYNITY, yet with much stronger reconstruction performance: this suggests the importance of correctly modeling both reaction and diffusion components of the model, even if one of them may be neural. 
These results 
provide evidence for the potential ability
of amortized-inference to learn across data samples.  

Note that
ALPS, with its \textit{perfect} mechanistic knowledge, achieved the best performance as expected among all baselines. 
Notably, 
HAPI-EP with its hybrid formulation 
(\textit{i.e.,} without perfect knowledge about the data generation mechanisms) achieved comparable performance without statistically significant difference to ALPS across all signal metrics except MAE for AT. 
More importantly, 
HAPI-EP obtained significantly improved identifiability results for the recovery of the physics-based and especially the neural latent parameters compared to all baselines, providing empirical evidence for its theoretical identifiability discussed in Section \ref{subsec:method:identifiability}.




The importance of this improved identifiability in the recovered latent parameters of the DT became evident in the prediction performances of all DTs in
the bottom half of 
Table~\ref{tab:main_results_ep_combined}.
While the relative performances among the baselines remained similar to those observed in reconstruction scenarios, 
all baselines' performance metrics deteriorated, 
highlighting the importance that a successful fitting to observed data does not guarantee a predictive DT. 
In comparison, 
HAPI-EP's performance remained remarkably stable across all metrics, 
delivering a more significant gain of performance over baselines in almost all metrics. 
Notably, 
a more detailed comparison between HAPI-EP and ALPS showed that ALPS was stronger in capturing the activation pattern, 
while HAPI-EP significantly outperformed ALPS in capturing 
recovery-related physiology, 
potentially owing to its significantly improved accuracy in recovering both latent diffusion and reaction parameters. 

This comparison between 
reconstruction and prediction performances in 
Table~\ref{tab:main_results_ep_combined} is important: 
the increasingly pronounced gap between HAPI-EP and its baselines in prediction (than reconstruction) 
supports its design premise: 
the importance of mechanistic backbones when data is weak, 
and the importance of amortized predictive-inference for transferable and identifiable recovery of latent parameters for a DT that is able to predict beyond observed data samples.

 \begin{table}[t]
\centering
\caption{Summary of out-of-distribution (OOD) quantitative results. 
Per-instance models are left out from this test because the OOD concept does not apply.
}
\label{tab:main_results_ep_ood}
\resizebox{\linewidth}{!}{
\begin{tabular}{lccc|ccc}
\toprule
\multirow[c]{2}{*}{\textbf{Method}} &
\multirow[c]{2}{*}{\textbf{MSE(OOD)}} &
\multirow[c]{2}{*}{\textbf{MSE(a)}} &
\multirow[c]{2}{*}{\textbf{MCC($\alpha$)}} &
\multicolumn{3}{c}{} \\
&&&& \textbf{MAE(APD)} & \textbf{MAE(AT)} & \textbf{MAE(RT)} \\
\midrule

ALPS \mTriS \cite{ALPS}                           & \textbf{0.020 $\pm$ 0.0223} & 0.0002 $\pm$ 0.0003 & 0.53 & 4.89 $\pm$ 4.39 & \textbf{0.73 $\pm$ 0.86} & 4.71 $\pm$ 4.30 \\

H-VAE \mTriH \cite{takeishi2021physvae}            & 0.0325 $\pm$ 0.0225 & 0.0002 $\pm$ 0.0003 & 0.52 & 5.9 $\pm$ 4.5 & 2.16 $\pm$ 1.27 & 4.99 $\pm$ 3.99 \\
HAPI-EP                                               & 0.0221 $\pm$ 0.009 & \textbf{3.91e-05 $\pm$ 3.69e-05} & \textbf{0.57} & \textbf{3.25 $\pm$ 1.6} & 1.99 $\pm$ 0.9 & \textbf{2.81 $\pm$ 1.52} \\
\bottomrule
\end{tabular}
}
\end{table}

\subsubsection{OOD Results}
Table~\ref{tab:main_results_ep_ood} reports the prediction performance of the three amortized models on held-out parameter settings where the reaction parameter takes intermediate values $a \in \{0.09, 0.11, 0.13\}$, which were not used during training, while the diffusion parameter remains within the same set $D \in \{1.5, 2.0, 2.5\}$ as in-distribution experiments. The per-instance models are omitted here because OOD is a irrelevant concept due to their training-free nature. 

As shown, 
compared to the hybrid H-VAE with amortized reconstructive-inference, 
HAPI-EP with its amortized predictive-inference showed significant improvements across all metrics. 
Even compared with ALPS based on perfect knowledge of the underlying data-generation mechanism, 
HAPI-EP still developed comparable MSE on the predicted action potential, better parameter identifiability, and improved estimation of APD and RT. 
Note that the improved MCC of $\alpha$ of ALPS and H-VAE compared to their in-distribution performance reported in Table \ref{tab:main_results_ep_combined} was because they performed poorly at higher values of $a=0.14$ which were not heavily presented in OOD tests.
These results provided evidence that the latent parameters HAPI-EP inferred were not only useful for interpolation, but remained  stable under distribution shift. 

\begin{figure}[t]
    \centering
    \includegraphics[width=1
    \linewidth]{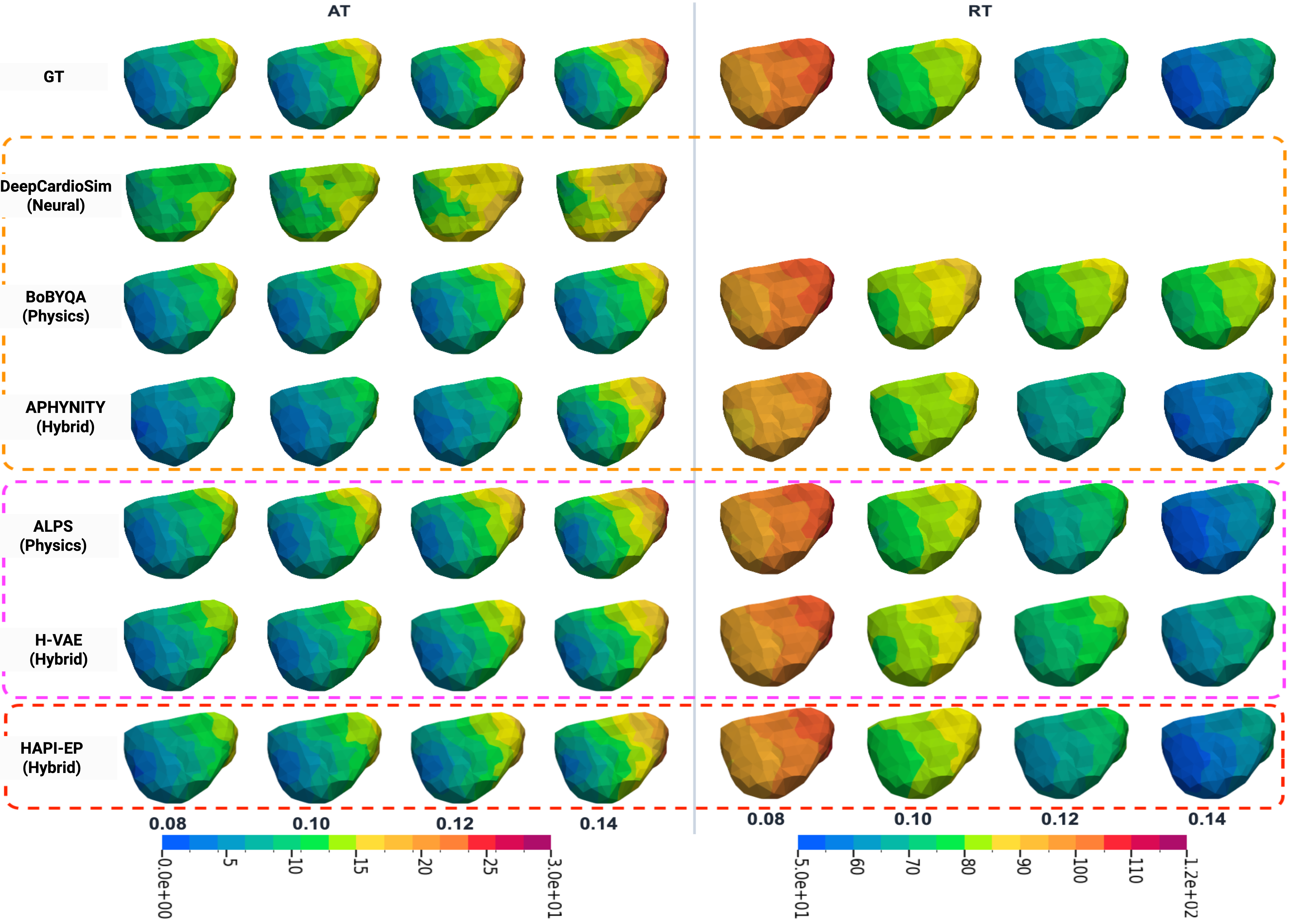}
     \caption{Examples of ground-truth (GT) \textit{versus} predicted activation time (AT) and repolarization time (RT) maps by all DTs, as the value of reaction parameter $a$ increased at a fixed $D = 1.5$. Orange box includes DTs optimized per-instance. Magenta box includes DTs optimized with amortized reconstructive-inference. 
     HAPI-EP in red box is optimized with amortized predictive-inference.
     }    \label{fig:a_var_maps}
    
\end{figure}
\subsubsection{Prediction Performance for Varying Reaction Parameters}

\begin{figure}[t]
\centering
\begin{subfigure}[t]{0.49\linewidth}
  \centering
  \includegraphics[width=\linewidth]{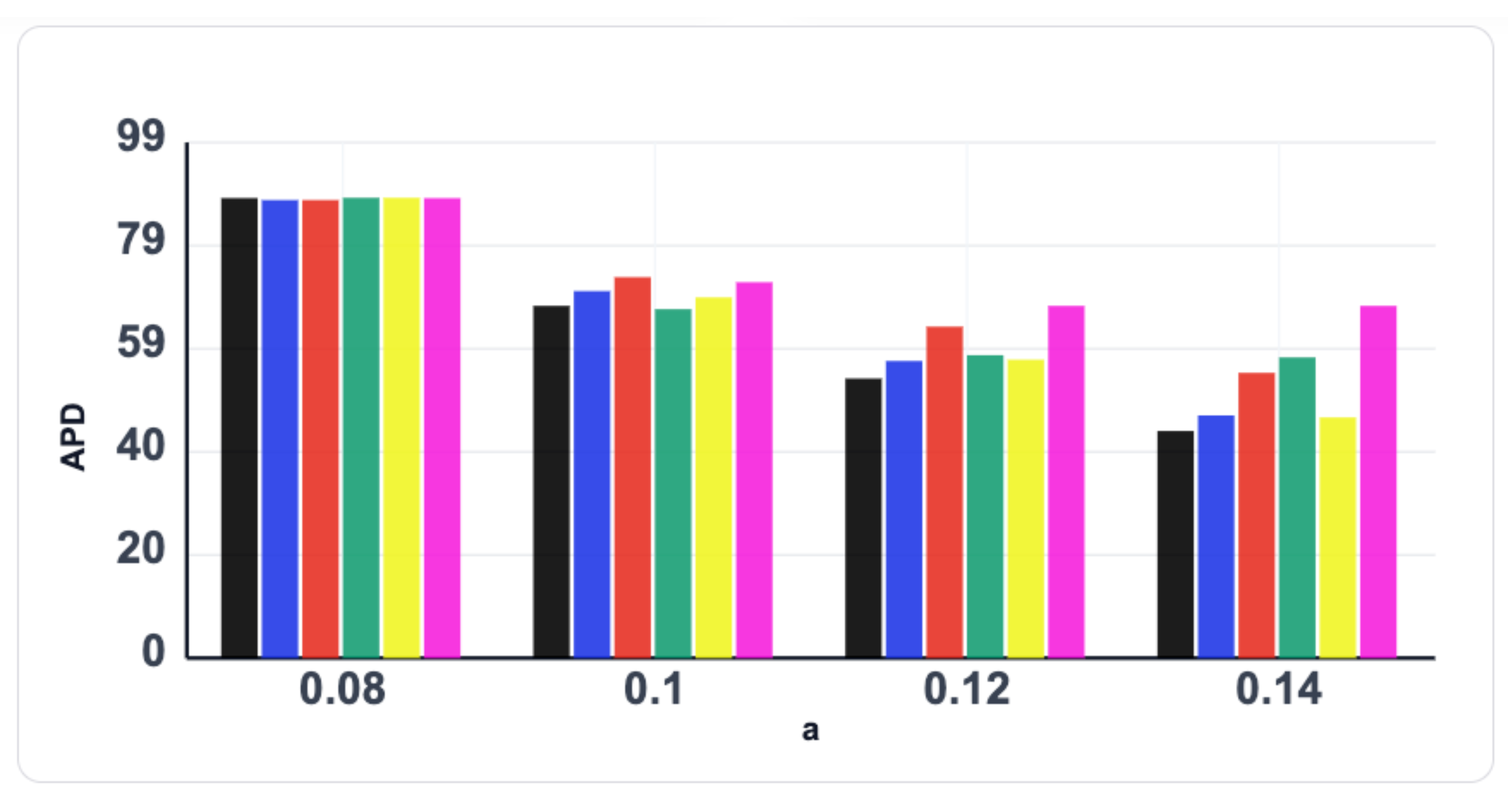}
  \caption{$D=1.5$}
\end{subfigure}\hfill
\begin{subfigure}[t]{0.49\linewidth}
  \centering
  \includegraphics[width=\linewidth]{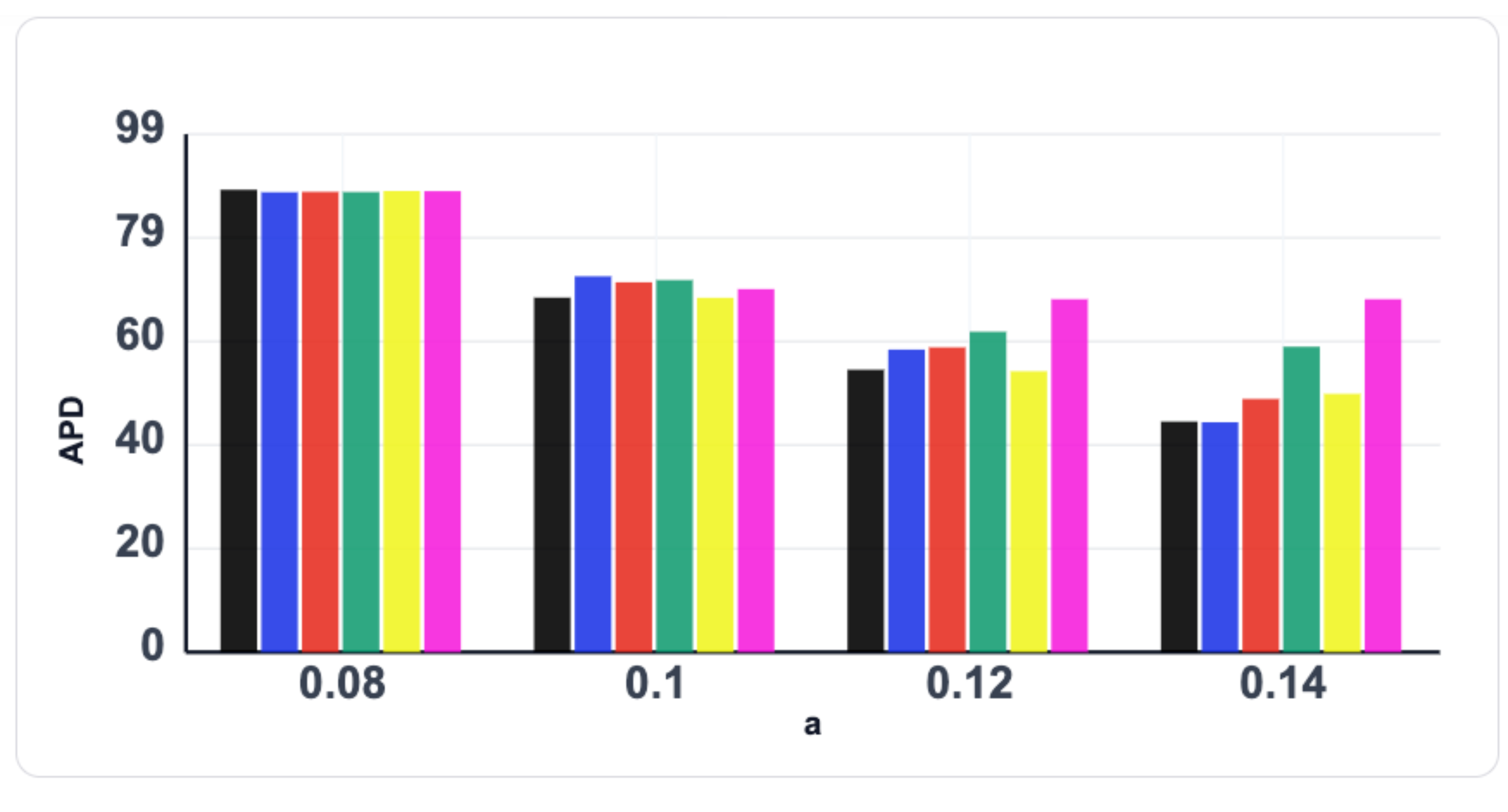}
  \caption{$D=2.0$}
\end{subfigure}

\vspace{2mm}

\begin{subfigure}[t]{0.49\linewidth}
  \centering
  \includegraphics[width=\linewidth]{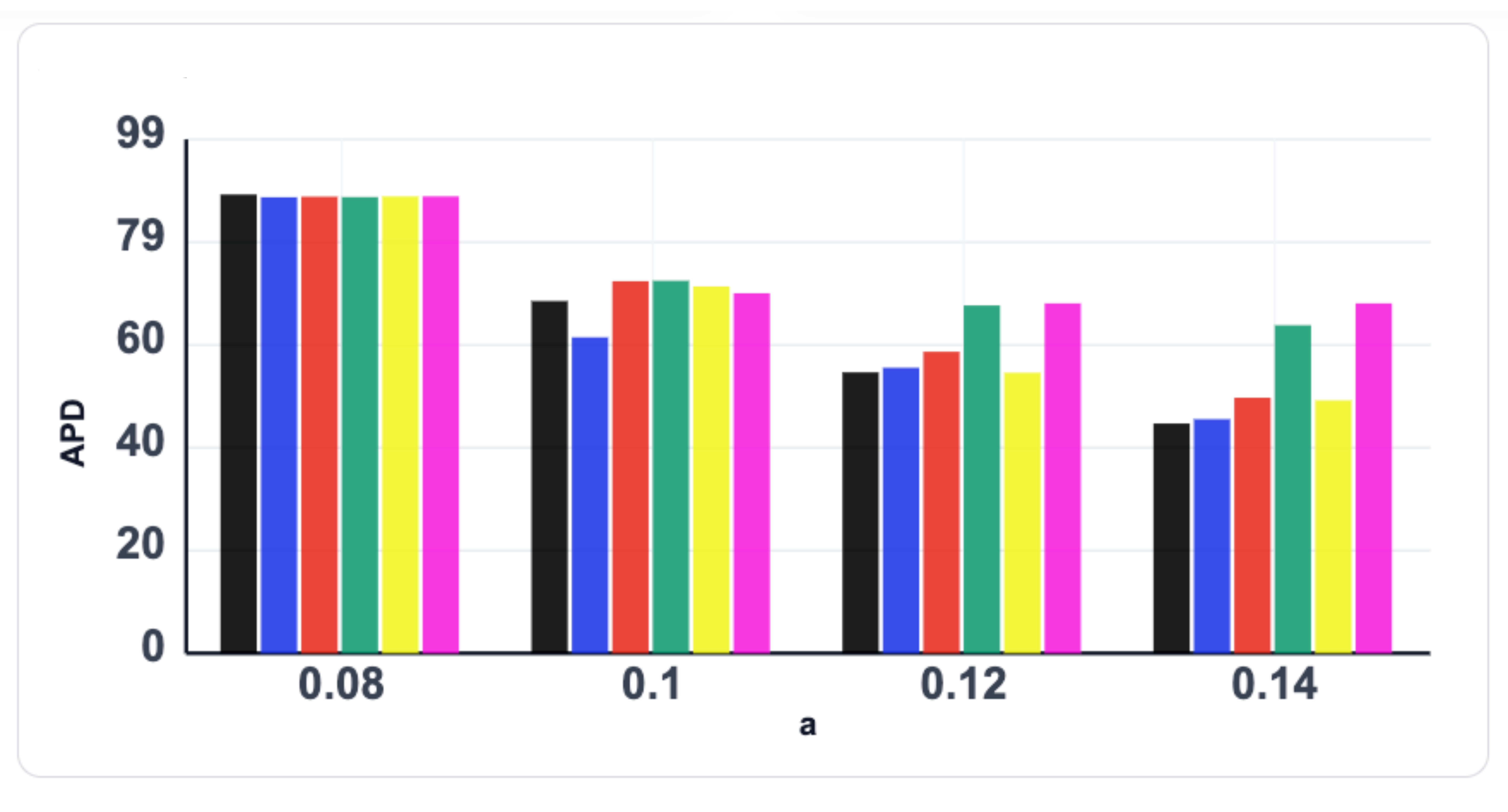}
  \caption{$D=2.5$}
\end{subfigure}\hfill

\vspace{2mm}
\includegraphics[width=0.5\linewidth]{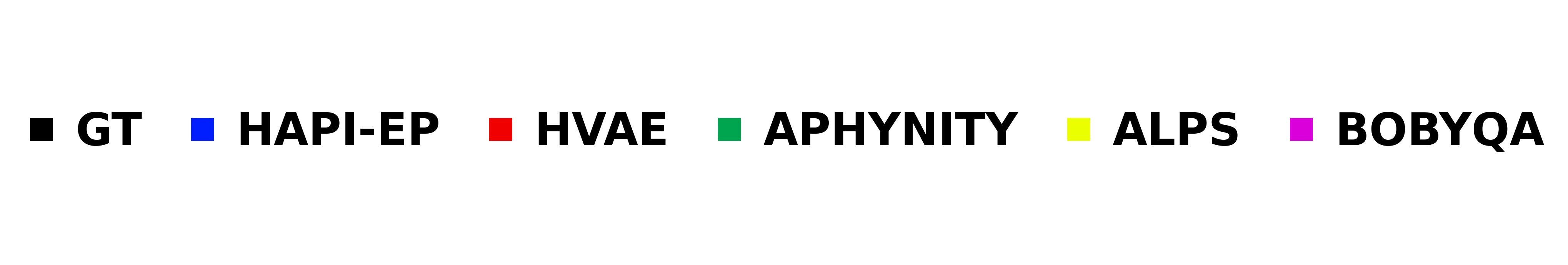}

\caption{Comparison of ground-truth (GT, black) and DT-predicted APD values across different values of reaction parameter $a$ at a given value of diffusion parameter $D$. }
\label{fig:apd_across_a_by_D}
\end{figure}

We now take a deeper dive into the prediction performance of all DTs under variations of the reaction parameter $a$. Figures~\ref{fig:a_var_maps} and~\ref{fig:apd_across_a_by_D} 
together demonstrate to what extent each DT preserves the spatial and physiological changes induced by $a$, using AT/RT maps and APD trends, respectively.

As shown in Figure~\ref{fig:a_var_maps}, in the synthetic ground-truth, the increasing value of parameter $a$ produced relatively modest increase in AT in local regions (\textit{e.g.,} basal-lateral right ventricular region), 
but much stronger overall reduction in RT across the ventricles, 
as expected by the intended effect of parameter $a$ on local action potential dynamics.

\textbf{Activation maps:}
In capturing the changes in AT, DeepCardioSim was able to reflect the overall trend of increasingly delayed activation as the value of $a$ increased. However, its predicted maps appeared less spatially coherent than the ground truth, with patchier transitions and a reduced contrast between early- and late-activation regions, particularly through the diminished blue early-activation area. This likely reflects the limitation of a purely neural surrogate without an explicit mechanistic model of propagation. 
The per-instance 
BoBYQA 
was able to perform well at lower parameter value of $a=0.08$ but not able to capture the increasingly delayed activation as $a$ increased; 
the per-instance APHYNITY, 
on the contrary, 
was able to capture the AT gradient at $a=0.14$ 
but reduced 
overall spatial gradient of AT maps at low and intermediate values of $a$: 
these performances may reveal the limitation of per-instance optimizations in leveraging the \textit{hidden} relations among samples, 
especially when contrasted with the performance of the three amortized DTs (ALPS, H-VAE, and HAPI-EP) which were all more successful in capturing the relative changes in AT maps as the value of $a$ increased.

Among the three amortized models, 
the reconstructive hybrid H-VAE exhibited a similar weakness to its per-instance counterpart (APHYNITY) in that it was less effective in capturing the contrast between the regions of early and late activation (\textit{i.e.}, the spatial gradient of AT maps) as $a$ increased. 
Both ALPS and HAPI-EP remained consistently close to the ground-truth across the full sweep in $a$, although it is worthwhile noting that ALPS' success was built on knowing the \textit{true} data-generating mechanisms in its DT models.

\textbf{Repolarization maps:}
In capturing the changes in RT, 
a similar trend in relative performances was observed among methods but at a more pronounced difference. 
Note that, as the value of $a$ increased, 
the ground-truth exhibited an overall reduction of RT throughout the ventricles.  
The per-instance mechanistic BoBYQA was faithful at $a=0.08$, but deteriorated substantially as the value of $a$ increased and clearly overestimated RT at higher values of $a$, particularly at $a=0.14$. 
The per-instance hybrid
APHYNITY captured the overall decrease in RT as $a$ increased, 
but reduced the spatial gradients of RT 
at all four $a$ values, elevating RT values
at regions of low-RT while reducing RT values at regions of high-RT on each map.
Among the three amortized models, H-VAE again was less effective in capturing the reduction of RT as the value of $a$ increased, especially at higher values of $a = 0.12$ and $0.14$.  
ALPS and HAPI-EP remained consistent and competitive across all values of $a$.

\textbf{Action potential duration:} 
Figure~\ref{fig:apd_across_a_by_D} provides a comprehensive summary of APD predictions of all methods against the ground-truth (black) as the value of parameter $a$ increased, at each fixed value of parameter $D$. 
Consistent with what was observed in Figures~\ref{fig:a_var_maps}, 
the ground-truth APD decreased consistently as the value of $a$ increased across all diffusion settings. 
The largest disagreement between all methods and the ground-truth appeared at $a=0.14$, matching the behavior seen in Figure~\ref{fig:a_var_maps}. 
BoBYQA showed increasing deviation from the truth as the value of $a$ increased, particularly at $a=0.12$ and $a=0.14$.
APHYNITY and H-VAE were reasonably close at lower value of $a$, but both showed larger departures from the ground-truth as $a$ increased, especially in the hardest regime of $a=0.14$. 
HAPI-EP and ALPS remained close to the decreasing trend in the ground-truth APD, across all diffusion settings.  
\begin{figure}[t]
    \centering
    \includegraphics[width=1
    \linewidth]{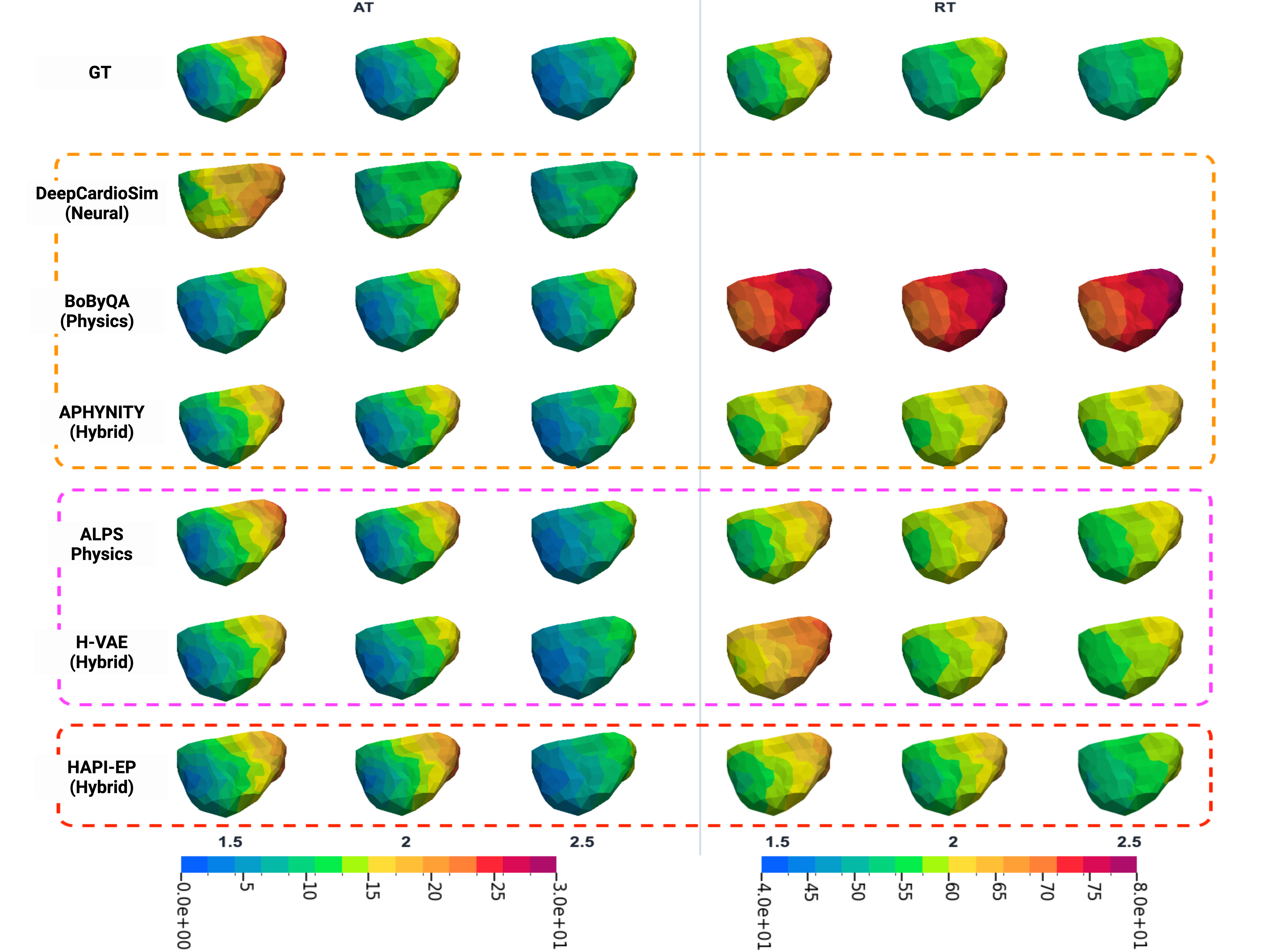}
     \caption{Examples of ground-truth (GT) \textit{versus} predicted activation time (AT) and repolarization time (RT) maps by all DTs, as the value of diffusion parameter $D$ increased at a fixed $a = 0.14$ Orange box includes DTs optimized per-instance. Magenta box includes DTs optimized with amortized reconstructive-inference. 
     HAPI-EP in red box is optimized with amortized predictive-inference.
     }
    \label{fig:d_var_maps}
\end{figure}
Taken together, Figs.~\ref{fig:a_var_maps} and~\ref{fig:apd_across_a_by_D} revealed a consistent pattern. Higher value of $a$, especially $a=0.14$, is the most difficult regime to recover. The effect of $a$ is more clearly expressed in RT and APD than in AT, making these outputs more diagnostic of whether a method has captured the underlying reaction-driven variation. 
The relative weaker performances of per-instance DTs in capturing the change in the underlying physiological variables as the value of $a$ changed revealed the benefit of learning across data samples in learning-based amortized inference. 

The strong performance of ALPS indicated the importance of mechanistic structure in improving stability, 
although it requires \textit{perfect} knowledge about the data-generating mechanisms. 
The competitive performance of HAPI-EP to ALPS, 
particularly in the high-$a$ regime where most other methods deteriorated, 
thus revealed its significant strength: 
its ability to resolve the residuals in imperfect prior mechanisms that was consistent across different underlying physiological conditions and 
at the same time is predictive across varying initial conditions.


\subsubsection{Prediction Performance for Varying Diffusion Parameters}

Figure~\ref{fig:d_var_maps} similarly shows AT and RT maps across increasing values of parameter $D$, at the fixed parameter value of $a=0.14$ --- 
the most difficult reaction regime observed earlier. 
As expected in the ground-truth, 
an increase value in the parameter $D$ increased the conduction velocity and thereby reduced 
the spread (\textit{i.e.,} spatial gradient) of both AT and RT maps across the ventricles.

\textbf{Activation maps:}
Similar to what was previously observed, DeepCardioSim was able to capture the overall trend of AT changes as $D$ increased, but its predicted maps appeared less spatially coherent than the ground truth. The per-instance BoBYQA 
was only able to capture the AT map at $D = 2.0$ without resolving their differences at different values of $D$, 
while similarly producing a similar RT map across different values of $D$ -- the absolute value of the RT maps was also incorrect, due to its inability to resolve the correct value of $a$ at $a = 0.14$ as revealed in Figure ~\ref{fig:a_var_maps}.  These combined results demonstrated the limitation of per-instance mechanistic BoBYQA in correctly resolving the change in both diffusion and reaction parameter values and how the difficulty in resolving one parameter may be affected by the other. 
The rest of the models were able to capture the change in AT maps as the value of $D$ increased, 
with the amortized hybrid H-VAE arguably being the most faithful to the ground-truth while the other three models (including HAPI-EP) showed an exaggerated spread of AT gradients for $D=2$.

\begin{figure}[t]
\centering

\begin{subfigure}[t]{0.49\linewidth}
  \centering
  \includegraphics[width=\linewidth]{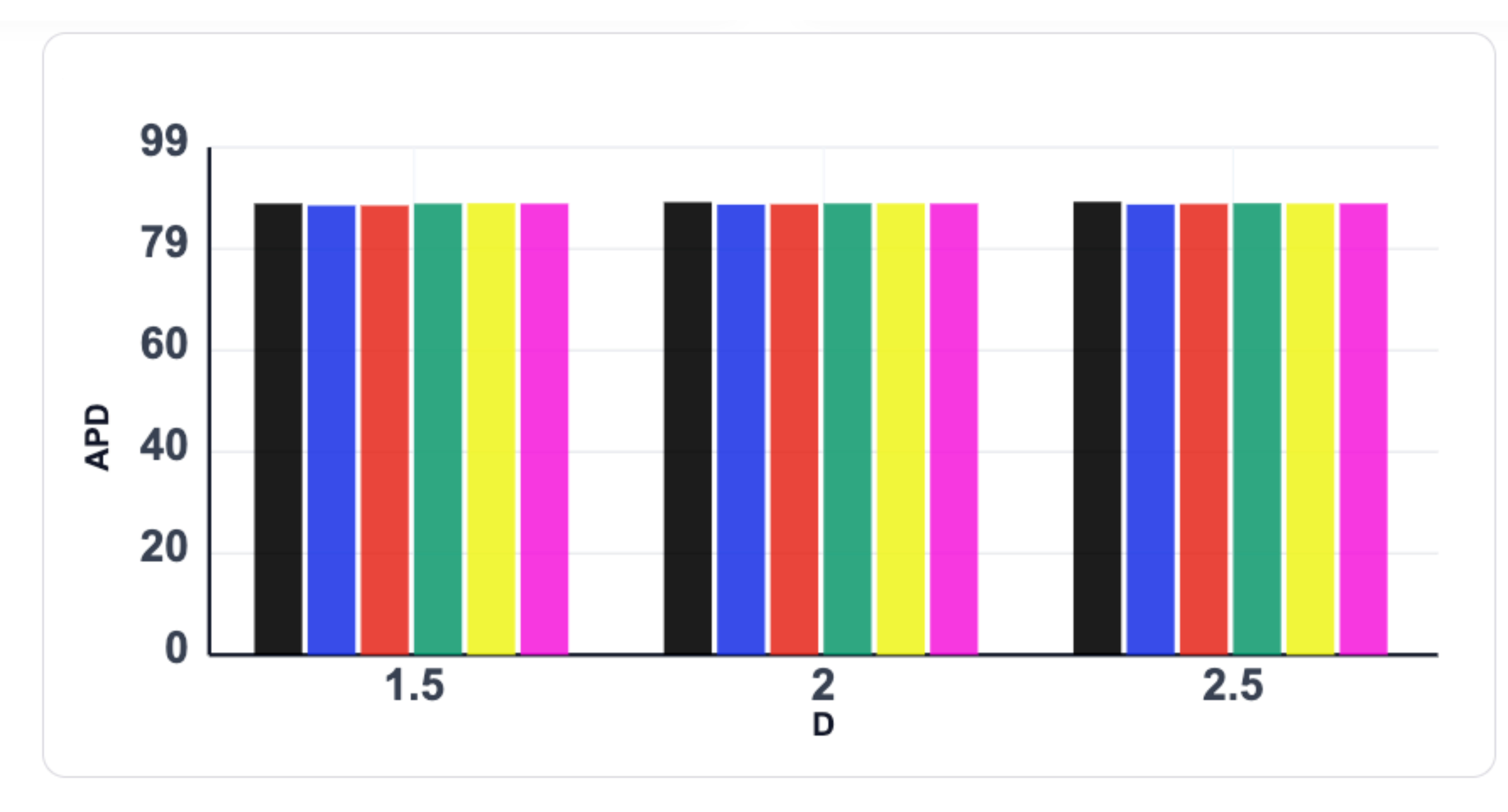}
  \caption{$a=0.08$}
\end{subfigure}\hfill
\begin{subfigure}[t]{0.49\linewidth}
  \centering
  \includegraphics[width=\linewidth]{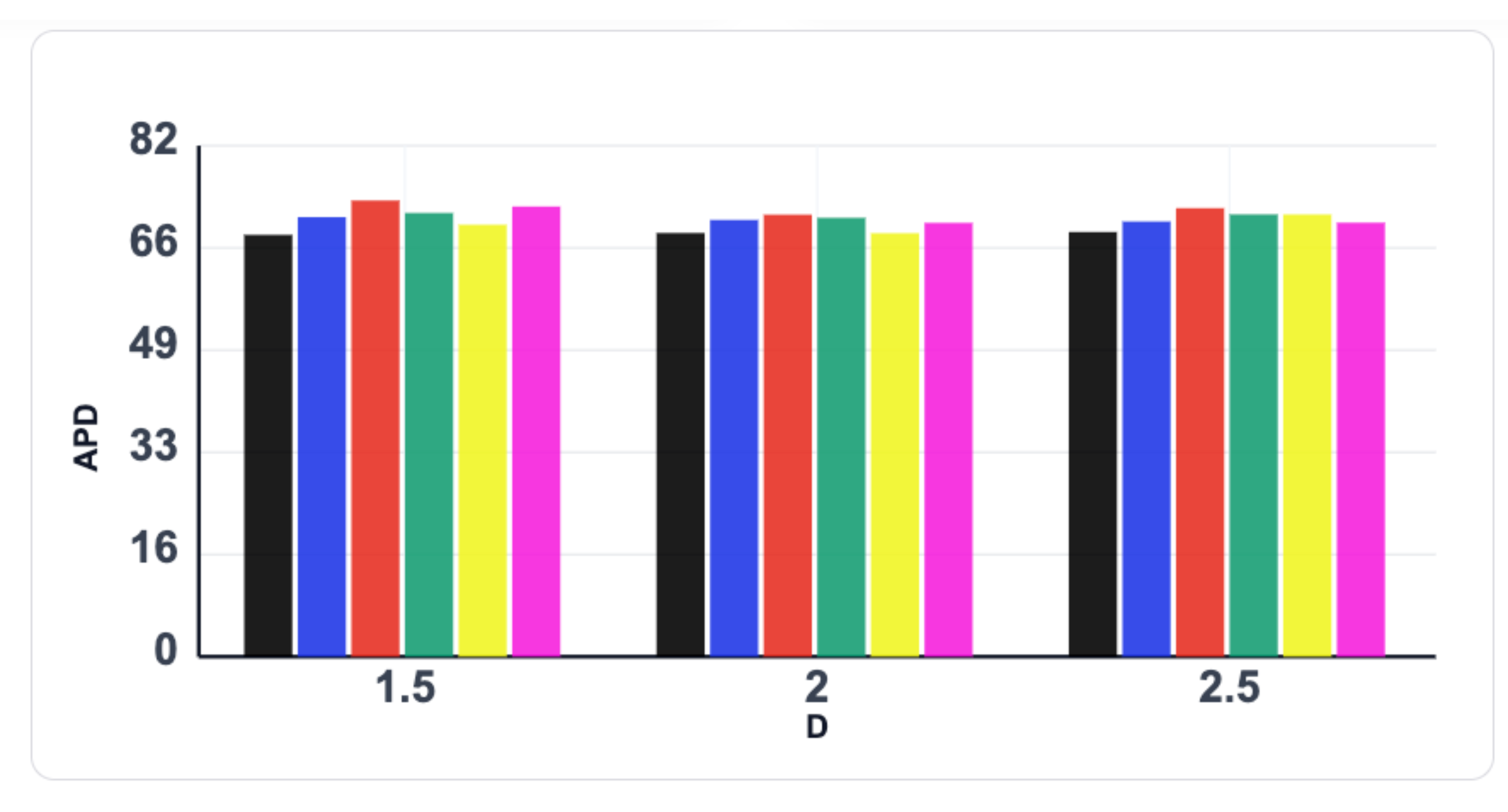}
  \caption{$a=0.10$}
\end{subfigure}

\vspace{2mm}

\begin{subfigure}[t]{0.49\linewidth}
  \centering
  \includegraphics[width=\linewidth]{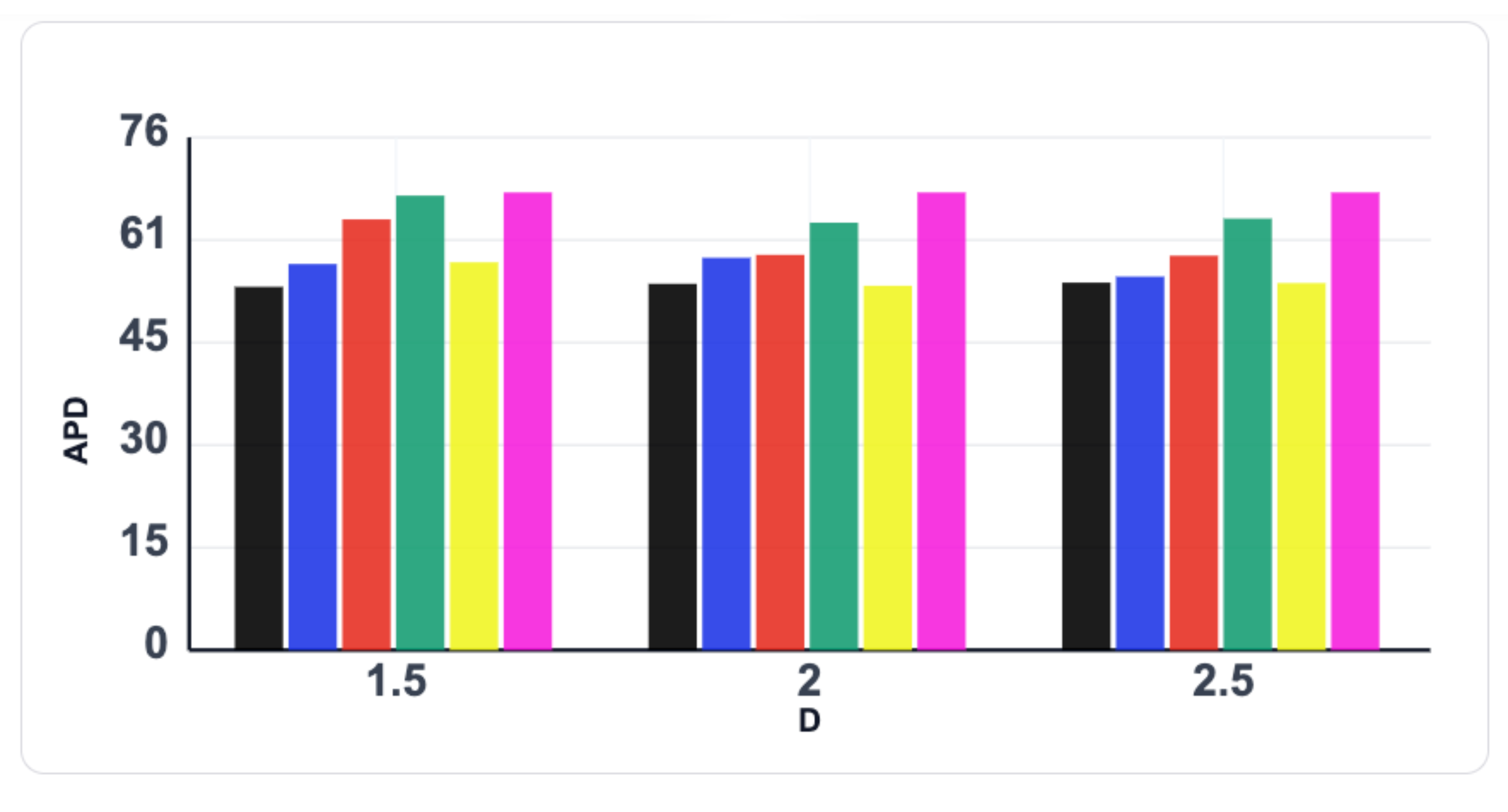}
  \caption{$a=0.12$}
\end{subfigure}\hfill
\begin{subfigure}[t]{0.49\linewidth}
  \centering
  \includegraphics[width=\linewidth]{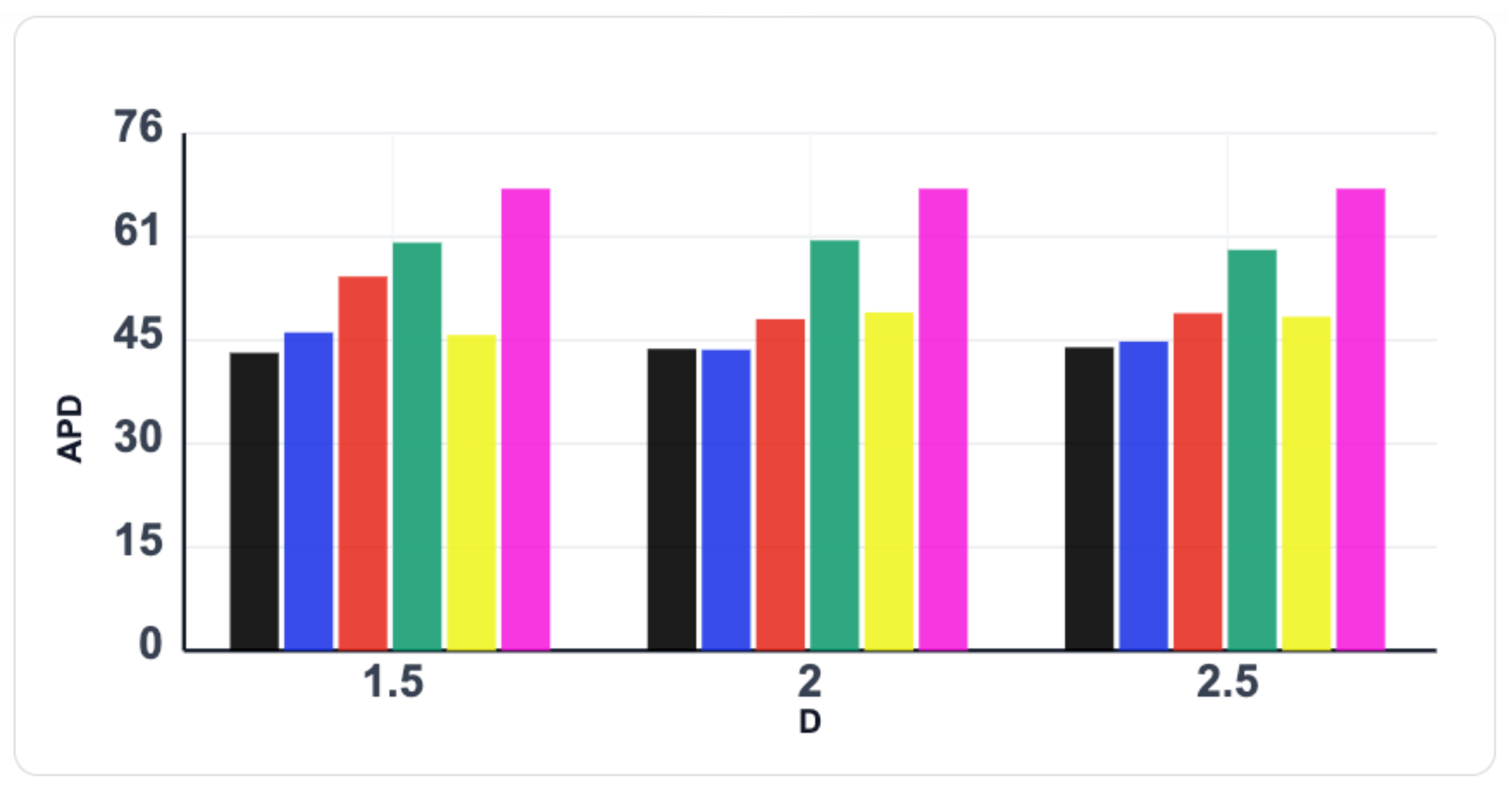}
  \caption{$a=0.14$}
\end{subfigure}

\vspace{2mm}
\includegraphics[width=0.5\linewidth]{doc/apd_legend_.png}

\caption{APD comparison (bar plots) across $D$ for different $a$ values. }
\label{fig:apd_bars_grid}
\end{figure}

\textbf{Repolarization maps:}
However, all baselines' prediction of RT maps were evidently less accurate than HAPI-EP. Per-instance hybrid APHYNITY, 
similar to previously observed, 
had a limited ability to resolve the difference in RT maps induced by the change of values in parameter $D$. 
The two amortized baselines were able to resolve this difference better, 
although the mechanistic ALPS 
showed an over-estimation of the spatial RT spread at higher values of $D=2.0$ and $2.5$, 
while the hybrid H-VAE over-estimated this spread throughout all values of $D$ with especially an over-estimate of the absolute RT values at lower values of $D=1.5$.
Among all models, 
HAPI-EP was the only DT that was able to consistently resolve the differences in AT and RT maps due to the different values of $D$.

\textbf{Action potential duration:} Figure~\ref{fig:apd_bars_grid} summarizes the predicted APD values by all DTs against the ground-truth (black) as the value of $D$ increased, at each fixed value of $a$. As expected, at a given parameter value of $a$, APD values should be minimally affected by the diffusion parameter $D$ as shown in the ground-truth (black). While all DTs were able to respect this at $a = 0.08$, all baselines started to increasingly over-estimate APD values as the value of $a$ increased, while HAPI-EP remained the most faithful to the ground-truth throughout all settings. Note that the amortized ALPS --- with \textit{perfect} mechanistic knowledge --- was competitive to HAPI-EP but still over-estimated at $a = 0.14$ compared to HAPI-EP. 

Taken together, Figs.~\ref{fig:d_var_maps} and~\ref{fig:apd_bars_grid} revealed a consistent pattern. Diffusion-driven physiological changes could be recovered reasonably well when the reaction regime is easy, but once $a$ became difficult to recover, the correct recovery of $D$ also became difficult. 
This highlights the challenge of parameter identification in DT adaptation which, without proper adaptation strategies,
could persist even with full knowledge about the data-generating mechanisms 
and became exaggerated if there were neural components within the DT. 
The \textit{learn-to-predict} strategies, 
underlined by theoretical identifiability of the latent DT parameters, 
were empirically proven to be the only strategy that was able to address this challenge in the experiments conducted. 

\begin{figure*}[t]
\centering

\begin{subfigure}[t]{0.32\linewidth}
  \centering
  \includegraphics[width=\linewidth]{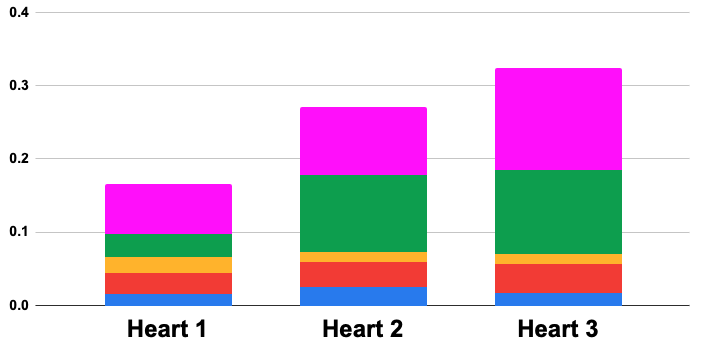}
  \caption{MSE(pred) across hearts.}
\end{subfigure}\hfill
\begin{subfigure}[t]{0.32\linewidth}
  \centering
  \includegraphics[width=\linewidth]{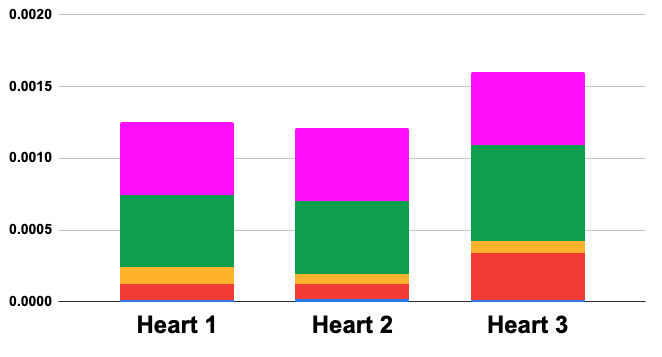}
  \caption{MSE(a) across hearts.}
\end{subfigure}\hfill
\begin{subfigure}[t]{0.32\linewidth}
  \centering
  \includegraphics[width=\linewidth]{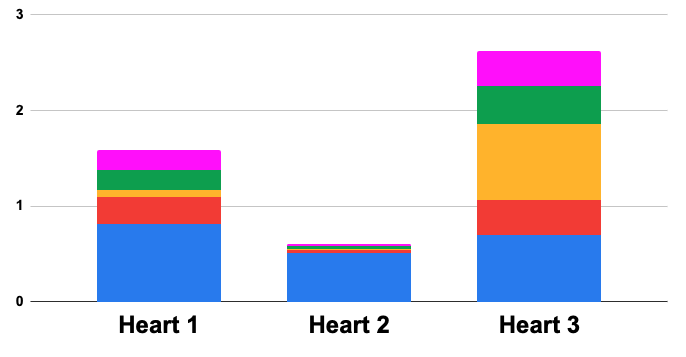}
  \caption{MCC($\alpha$) across hearts.}
\end{subfigure}

\vspace{2mm}

\begin{subfigure}[t]{0.49\linewidth}
  \centering
  \includegraphics[width=\linewidth]{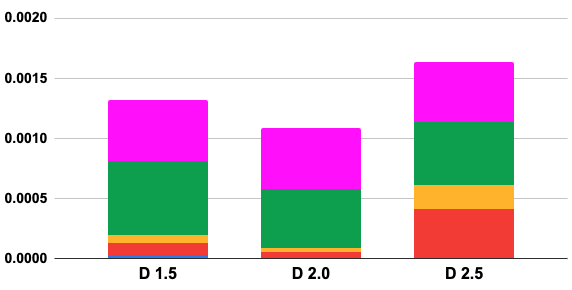}
  \caption{MSE(a) across $D$.}
\end{subfigure}\hfill
\begin{subfigure}[t]{0.49\linewidth}
  \centering
  \includegraphics[width=\linewidth]{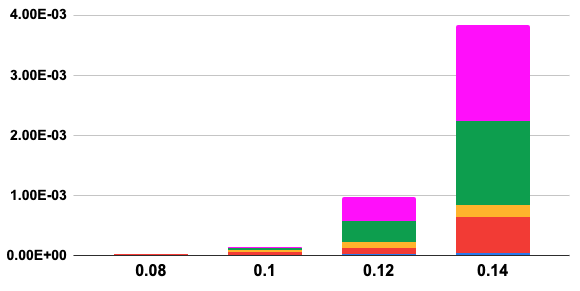}
  \caption{MSE(a) across $a$.}
\end{subfigure}

\vspace{2mm}
\includegraphics[width=0.5\linewidth]{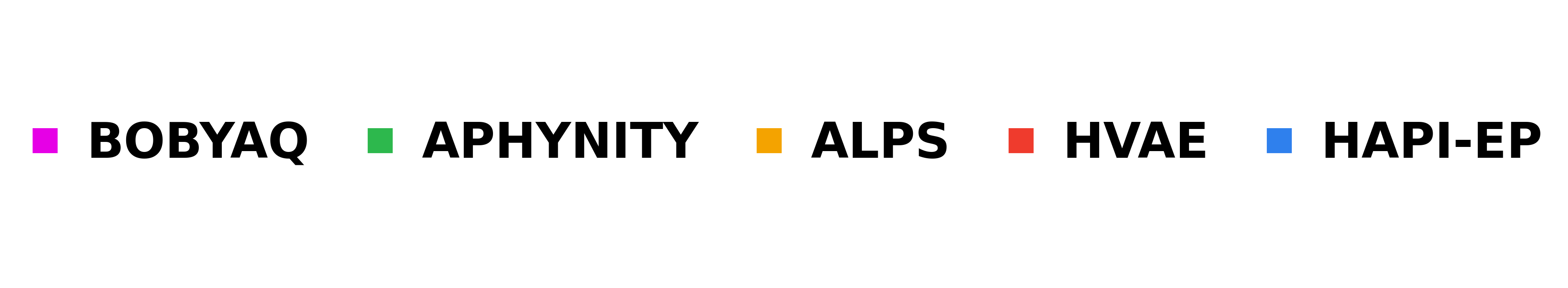}

\caption{Summary metrics (stacked bar plots) across hearts and parameter settings.
}
\label{fig:stacked_metrics_grid}
\end{figure*}

\subsubsection{Quantitative Comparisons across Hearts and Physiological Conditions}

Figure~\ref{fig:stacked_metrics_grid} summarizes these quantitative trends across hearts and conditions, 
to examine the stability of each DT across hearts/conditions as well as identify the most challenging scenarios for all DTs.

Across the three ventricles considered, 
as shown in Figure~\ref{fig:stacked_metrics_grid}(a)-(c), 
the identification of reaction parameter $a$ seemed to be the most difficult in heart 3, while the identification of diffusion parameter $D$ was significantly the most difficult in heart 2.  
Across the physiological values of parameters $a$ and $D$,
Figure~\ref{fig:stacked_metrics_grid}(d)-(e) show how 
the largest error was seen at
$D=2.5$ 
while 
the increase in the value of $a$ significantly increased the difficulty.


Within each setting, the stacked bars also show the extent to which each DT was affected by the difficulty of estimation.
Overall, across all settings, 
the largest contributions to prediction and parameter-estimation errors tended to come from BoBYQA and APHYNITY (tall MSE bars and short MCC bars),  
particularly in the more difficult settings, 
whereas HAPI-EP consistently contributed the lowest errors 
(short MSE bars and tall MCC bars) 
across the hearts and physiological conditions considered.

\begin{table}[t]
\centering
\caption{Additional ablation of the two key elements in HAPI-EP. The shaded column represents a best-case scenario because it uses a known mechanism.}
\label{tab:ablation_horizontal}
\setlength{\tabcolsep}{10pt}
\renewcommand{\arraystretch}{1.15}
\begin{tabular}{lcccc>{\columncolor{gray!15}}c}
\hline
\textbf{Hybrid} & X & \checkmark & X(neural) & \checkmark & X(physics) \\
\textbf{Learn to predict} & X & X & \checkmark & \checkmark & \checkmark \\
\textbf{MSE}($\mathbf{x}$) & 0.193 & 0.072 & 1.3 & 0.019 & 0.012 \\
\hline
\end{tabular}
\end{table}

\subsubsection{Additional Ablations} 
\label{subsec:ablations}

We conducted additional ablation studies on HAPI-EP, progressively testing each of its key elements: the hybrid formulation, and its identifiable formulation with the amortized predictive-inference.  
As summarized in
Table \ref{tab:ablation_horizontal},
The starting ablation, 
without the hybrid component or the 
\textit{learn-to-predict} objective, 
was a neural graph diffusion architecture the same as that used in HAPI-EP but with a neural function for the reaction terms, 
optimized per instance.
Adding the amortized predictive-inference along, 
using the same encoder architectures as described in HAPI-EP, 
was not successful, 
showing the importance of mechanistic knowledge in this DT. 
Indeed, 
adding the hybrid structure alone 
improved the performance substantially, 
although still to a limited extent. 
The use of hybrid formulation together with the \textit{learn-to-predict} objective significantly improved the performance of the DT, to a level that is competitive to the use of the \textit{perfect} data-generating mechanism (shaded).  
These results reiterated that both the use of a hybrid model formulation and the amortized predictive-inference strategies are necessary for an adaptive and predictive DT. 

\section{Real Data Experiments}

\begin{figure}[t]
    \centering
    \includegraphics[width=1
    \linewidth]{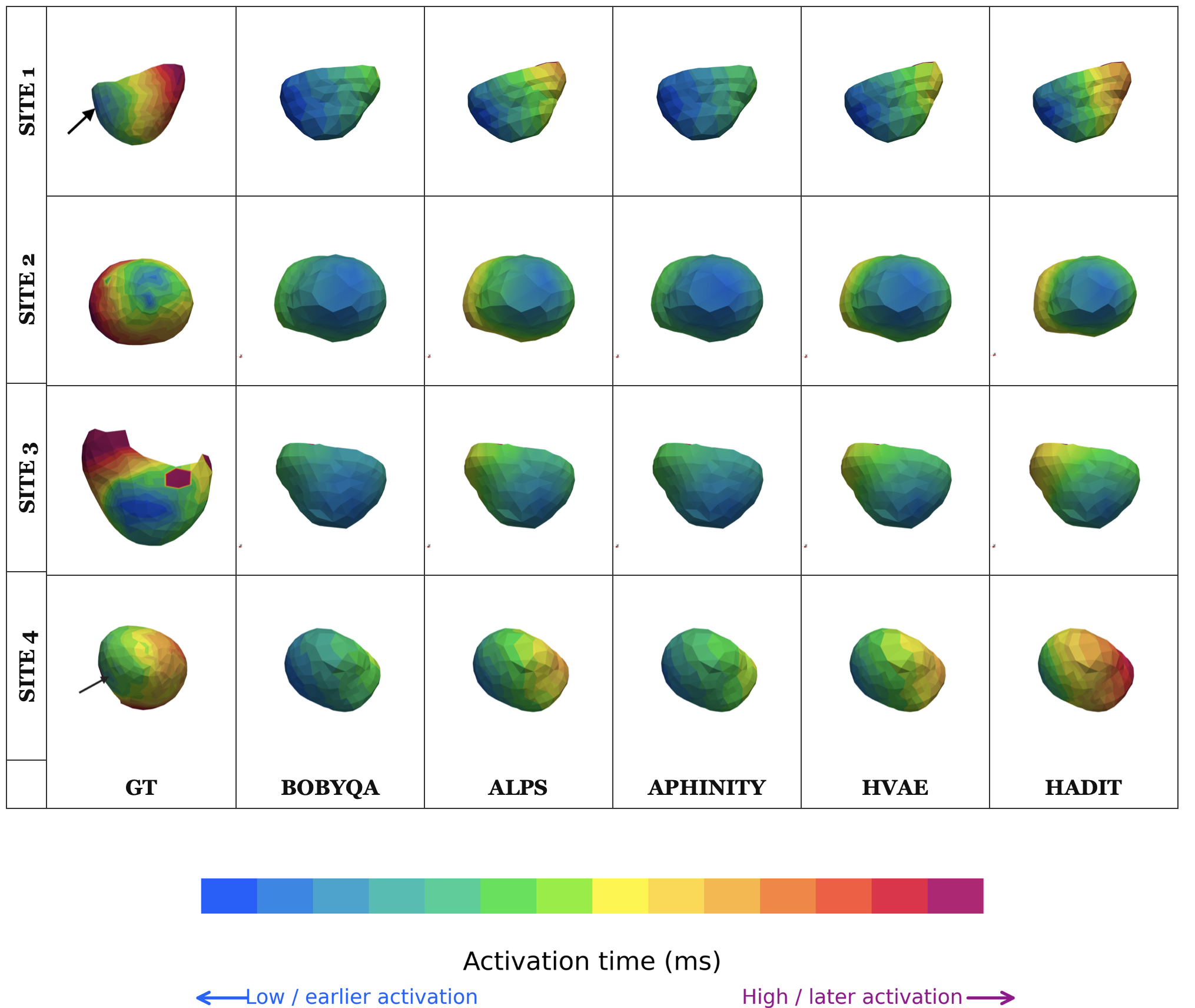}
     \caption{Visual comparison of predicted AT maps for four pacing sites by all models (low$\rightarrow$high)}
    \label{fig:real_at_maps}
\end{figure}
\begin{figure}[t]
    \centering
    \includegraphics[width=0.9
    \linewidth]{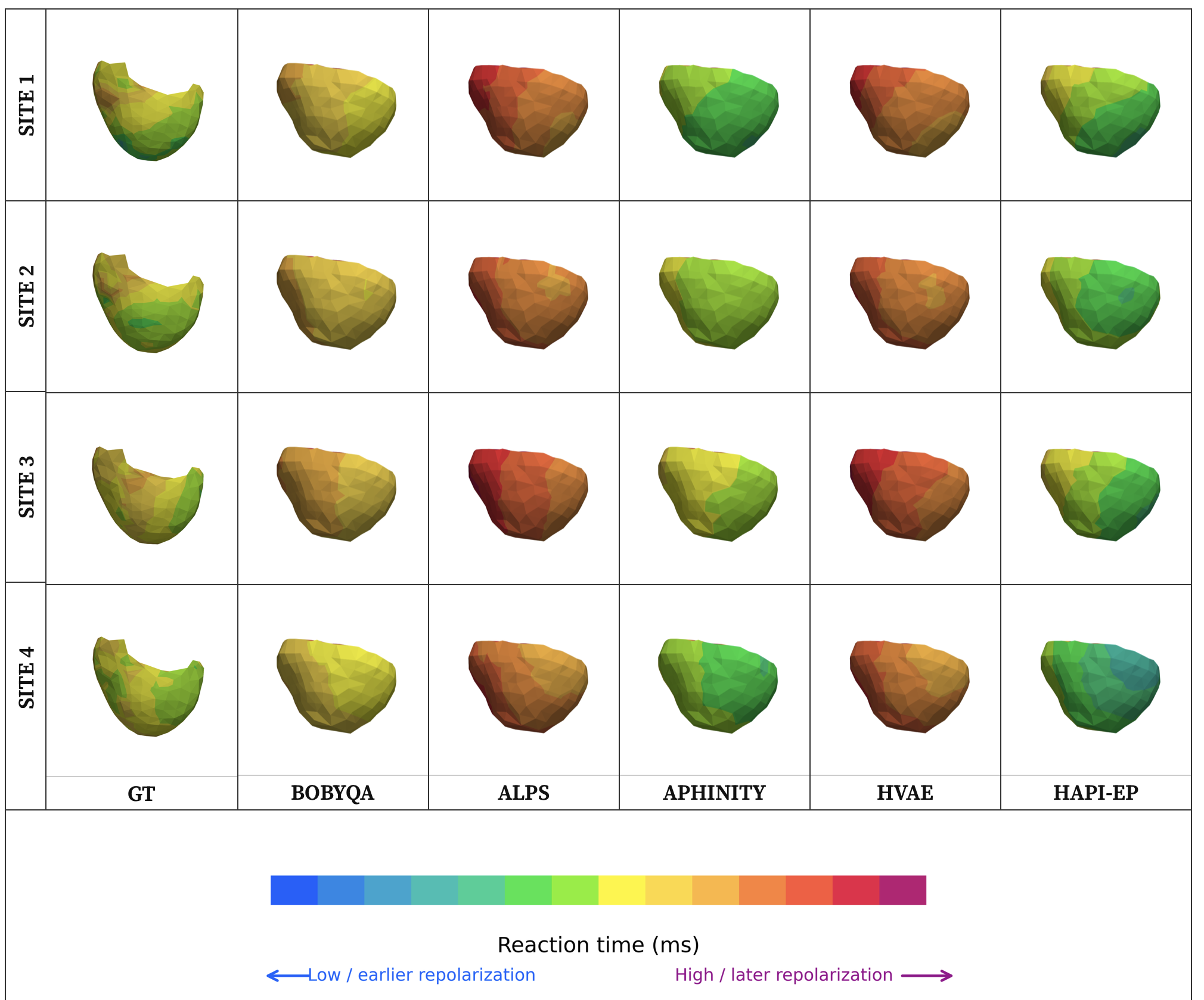}
     \caption{Visual comparison of predicted RT maps for four pacing sites by all models (low$\rightarrow$high).}
    \label{fig:real_rt_maps}
\end{figure}

\begin{table}[t]
\centering
\caption{Summary of quantitative results and computation time (wall-clock time in seconds for  adaptation and prediction) on real-data.}
\label{tab:ep_results}
\resizebox{\linewidth}{!}{
\begin{tabular}{lcccc}
\toprule
\textbf{Method} & \textbf{MAE(AT)} & \textbf{MAE(APD)} & \textbf{MAE(RT)} & \textbf{Computation (s)} \\
\midrule
DeepCardioSim \mSqrH   & 6.39 $\pm$ 1.29 & -- & -- & 6.39 \\
BoBYQA \mSqrH  & 10.18 $\pm$ 2.45 & 15.79 $\pm$ 4.04 & 12.22 $\pm$ 1.53 &  1630.56 \\
APHYNITY \mSqrH & 9.61 $\pm$ 2.80  & 10.88 $\pm$ 0.96 & \textbf{9.29 $\pm$ 2.39} &3293.96 \\
ALPS \mTriS    & 6.55 $\pm$ 2.27  & 22.21 $\pm$ 4.51 & 16.12 $\pm$ 2.67 & 30.13 \\

H-VAE \mTriS   & 6.73 $\pm$ 2.65  & 21.95 $\pm$ 4.55 & 15.64 $\pm$ 2.49 & \textbf{28.20} \\
HAPI-EP   & \textbf{5.14 $\pm$ 1.96} & \textbf{9.63 $\pm$ 4.31} & 11.38 $\pm$ 3.66 & 32.24 \\
\bottomrule
\end{tabular}
}
\end{table}

\subsection{Experimental Settings}

We evaluate on \textit{in-vivo} canine epicardial recordings acquired with a multi-electrode epicardial sock under controlled pacing protocols, following the experimental setup and preprocessing pipeline described in prior work~\cite{bergquist2021electrocardiographic}.
Bipolar plunge-needle pacing was delivered at 4 unique left-ventricular sites to generate activation sequences, and EGMs were recorded at 1\,kHz on an sock array applied to the epicardium only.

We considered all models from Section \ref{subsec:exp:synth_data}, adapting the DT using real EGM data for \textit{predictive scenarios:} 
in a cross-validation setting, 
we used epicardial EGMs from three pacing sites to infer 
latent DT model parameters, and the adapted model was used to make predictions for the one left-out pacing site; this was done four times, leaving recordings from one pacing site out each. For DeepCardioSim which as a neural surrogate outputs only AT, per-instance optimization was carried out using EGM-derived AT data.
For the base DeepCardioSim network and the amortized models including HAPI-EP, since there was not sufficient \textit{in-vivo} data for training, 
the models trained on the synthetic data in Section \ref{subsec:exp:synth_data} were directly tested here.

For evaluation, we extracted AT and APD from the predicted action potential signals, and evaluated their MAEs against AT and APD derived from EGMs measured on the epicardial sock.

\subsection{Results}

Figure~\ref{fig:real_at_maps} compares AT maps predicted by all DTs across the four pacing sites. The ground-truth maps show clear site-dependent changes in both the location of the high-AT region and the overall spatial gradient. 
The relative performance of all DTs resembled that observed in synthetic experiments. 
DeepCardioSim was able to predict the general early-to-late AT pattern, although consistently under-performed in the region near the site of early activation. 
The per-instance BoBYQA and APHYNITY were able to capture 
an activation pattern starting from each pacing site, 
although both under-estimated the spatial gradient of the AT maps across all four cases. 
The two amortized reconstructive DTs, ALPS and HVAE, 
were more successful in capturing the AT gradients in space, although still under-estimated the complete span of AT exhibiting smaller latest activation time in all four cases.  
HAPI-EP predicted the most consistent AT maps to the ground-truth in all cases, preserving both the site-specific region of early- and late-activation regions as well as the range of AT values. 

Figure~\ref{fig:real_rt_maps} compares RT maps across the same pacing sites. In contrast to AT, the measured RT maps were more similar across sites in their global structure, with subtle site-dependent changes in spatial distributions. 
Interestingly, the relative performance of the DTs was different from what was observed in synthetic data. 
Notably, the two mechanistic DTs --- BoBYQA and ALPS --- significantly over-estimated RT values. 
This suggested that mechanistic DTs, 
while performing well when \textit{perfect} mechanistic knowledge is available (\textit{e.g.,} in synthetic data), 
can deteriorate significantly when the underlying mechanistic knowledge has a gap to real data. 
Between the two hybrid DTs that had learnable neural component to resolve this gap, 
the amortized H-VAE also had a similar problem of over-estimating the RT ranges, 
while the per-instance APHYNITY performed significantly better. 
This suggested the challenge of generalization associated with learning-based amortized inference, especially when training occurred on synthetic data. 
Notably, the learning-based HAPI-EP was not impacted by this issue and performed visually similar to APHYNITY in its predicted RT maps across all four sites. 
This highlighted the benefit of its \textit{learn-to-predict} objective.

Quantitative MAE values in Table~\ref{tab:ep_results} confirmed these visual comparisons. HAPI-EP achieved the lowest MAE in AT, with DeepCardioSim being the next best method on this metric. Notably, however, DeepCardioSim was personalized directly from AT information, whereas the DT baselines adapted from EGM measurements, making its setting narrower and more supervised. 
APHYNITY achieved the lowest MAE in RT, although its difference to HAPI-EP was within one standard deviation of the mean. Importantly, HAPI-EP (and ALPS) also achieved the lowest MAE in the predicted APD values, which was substantially improved over the rest of the baseline DTs.
Importantly, 
in the last column of Table~\ref{tab:ep_results} we show 
the computation (wall-clock time in seconds) for each method in adapting the DT and using the adapted DT for prediction, measured per case with the average reported across all cases.
As shown, 
although HAPI-EP and APHYNITY achieved similar performance in RT and APD predictions, 
HAPI-EP was able to deliver this adaptation at a fraction of the computation cost of the per-instance APHYNITY.

\section{Conclusions and Discussion}

In this paper, 
we present HAPI-EP as a novel foundation for building hybrid and adaptive DTs 
with predictive-ness guaranteed by the identifiability of its latent parameters. 
Our synthetic and real-data experimental results showed several important findings. 
First, 
activation and repolarization features were not equally easy to recover, and the latter 
seemed to present a bigger challenge for most per-instance based approaches whether the underlying model is mechanistic or not. 
In this context,
amortized inference, 
\textit{i.e.}, learning the optimization across many instances, 
seemed to offer not only benefits of rapid adaptation at deployment time but also improvements in resolving repolarization-related features from data.
Second, 
the performance of parameter adaptation of a method can change depending on the values of the parameter, and this 
 difficulty can impact the method's ability to resolve another parameter.
For instance, 
$a=0.08$ appeared the easiest for all DTs to estimate correctly, 
while a larger performance gap was observed among methods at high values of parameter $a$ (Fig.~\ref{fig:apd_across_a_by_D}) across all values of parameter $D$ 
(Fig.~\ref{fig:apd_bars_grid}). Finally, 
HAPI-EP with its hybrid formulation and amortized predictive-inference consistently provided the strongest performance across parameter values, especially in improving predictive accuracy and OOD performance.








As a proof-of-concept, 
this work has the following limitations that can be addressed in future investigations. 
In terms of formulation of the hybrid DT, 
while the concept and methodology of HAPI-EP is not tied to the specific choice of mechanistic models, 
we demonstrated its proof of concept using the reaction mechanism described in the Aliev-Panfilov model combined with a neural graph diffusion.
A natural next step is to extend HAPI-EP to richer ionic models and assess whether its hybrid and identifiable adaptation strategy remains effective when the mechanistic component is more biophysically detailed. 
Other more general forms of hybridization between mechanistic and neural components can also be considered in future works.  
In terms of the inference methodology, 
the current study is limited to the 
setting where the activation site is controlled/known: 
when the initial site of activation is not known, 
this will become an additional learnable latent variable whose identifiability needs to be established along with the latent parameters of the hybrid model. 
Furthermore,
for simplicity, 
the latent parameters $\mathbf{z}_\text{PHY}$ and $\mathbf{z}_\text{NN}$ were assumed to be Gaussian with fixed variance parameter $\sigma$. While this stabilizes training, it limits the flexibility of the uncertainty model. A more expressive formulation that learns or adapts this variance may better capture heterogeneity in confidence across tasks and subjects. 
Finally, 
although HAPI-EP enables rapid feedforward adaptation at test time, this comes at a nontrivial computation and memory cost during training because the framework jointly optimizes context encoders, hypernetworks, and the hybrid simulator in an episodic meta-learning setting. Reducing this training-time overhead will be important for scaling the method to larger models and datasets.

In experimental evaluation of HAPI-EP, 
the synthetic training data were generated from the fully mechanistic Aliev-Panfilov model. 
While this allowed a controlled testing of whether the hybrid formulation of HAPI-EP is able to resolve the knowledge gap about the mechanistic diffusion, 
it does not fully capture the structural mismatch encountered in practice. Future work should therefore evaluate HAPI-EP under higher-fidelity simulation environments, including more detailed electrophysiology models and more realistic forward models, to better test the robustness of the hybrid formulation to capture mismatch in prior mechanistic knowledge.
Furthermore, 
while we considered the heterogeneity of the reaction and diffusion parameters in our experiments, 
these parameters were assumed to be spatially uniform. 
How to take into account spatially-varying 
electrophysiological properties is an important future direction in order to model local pathological conditions, 
including non-trivial challenges such as establishing conditions that need to be met in order to meet the identifiability of such high-dimensional parameters  .

Finally, while the \textit{in-vivo} evaluation provided promising evidence for the generalizability of HAPI-EP even when trained on simulation data, it remained limited in scale. Larger studies on both animal and human data will be necessary to establish the robustness, clinical relevance, and generalizability of HAPI-EP across geometries, physiological conditions, and measurement conditions.

\section*{Acknowledgement} 
This work was supported by National Institutes of Health (NIH) award number
R01HL145590 (contact PI Wang).

\bibliography{miccai2025}
\bibliographystyle{splncs04}
\end{document}